\documentclass{article}

\usepackage{microtype}
\usepackage{graphicx}
\usepackage{subfigure}
\usepackage{booktabs} 

\usepackage{algorithm} 
\usepackage{algpseudocode} 
\usepackage{color, soul} 
\usepackage{multirow} 

\usepackage{hyperref}

\usepackage[accepted]{icml2025}

\usepackage{amsmath}
\usepackage{amssymb}
\usepackage{mathtools}
\usepackage{amsthm}

\usepackage[capitalize,noabbrev]{cleveref}

\theoremstyle{plain}

\theoremstyle{definition}

\theoremstyle{remark}

\usepackage[textsize=tiny]{todonotes}

\begin{document}

\twocolumn[
\icmltitle{No More Adam: Learning Rate Scaling at Initialization is All You Need}



\icmlsetsymbol{equal}{*}

\begin{icmlauthorlist}
\icmlauthor{Minghao Xu}{equal,sch}
\icmlauthor{Lichuan Xiang}{equal,sch,comp}
\icmlauthor{Xu Cai}{comp}
\icmlauthor{Hongkai Wen}{sch}
\end{icmlauthorlist}

\icmlaffiliation{sch}{Department of Computer Science, University of Warwick, Coventry, UK}
\icmlaffiliation{comp}{Collov Labs}

\icmlcorrespondingauthor{Hongkai Wen}{hongkai.wen@warwick.ac.uk}
\icmlkeywords{Machine Learning, ICML}

\vskip 0.3in
]



\printAffiliationsAndNotice{\icmlEqualContribution} 

\begin{abstract}

In this work, we question the necessity of adaptive gradient methods for training deep neural networks. SGD-SaI is a simple yet effective enhancement to stochastic gradient descent with momentum (SGDM). SGD-SaI performs learning rate Scaling at Initialization (SaI) to distinct parameter groups, guided by their respective gradient signal-to-noise ratios (g-SNR). By adjusting learning rates without relying on adaptive second-order momentum, SGD-SaI helps prevent training imbalances from the very first iteration and cuts the optimizer’s memory usage by half compared to AdamW. Despite its simplicity and efficiency, SGD-SaI consistently matches or outperforms AdamW in training a variety of Transformer-based tasks, effectively overcoming a long-standing challenge of using SGD for training Transformers. SGD-SaI excels in ImageNet-1K classification with Vision Transformers(ViT) and GPT-2 pretraining for large language models (LLMs, transformer decoder-only), demonstrating robustness to hyperparameter variations and practicality for diverse applications. We further tested its robustness on tasks like LoRA fine-tuning for LLMs and diffusion models, where it consistently outperforms state-of-the-art optimizers. From a memory efficiency perspective, SGD-SaI achieves substantial memory savings for optimizer states, reducing memory usage by 5.93 GB for GPT-2 (1.5B parameters) and 25.15 GB for Llama2-7B compared to AdamW in full-precision training settings. \footnote{The PyTorch implementation is available at \url{https://github.com/AnonymousAlethiometer/SGD_SaI/}}
\end{abstract}    

\begin{figure}[h!]
    \centering
    \includegraphics[width=\columnwidth]{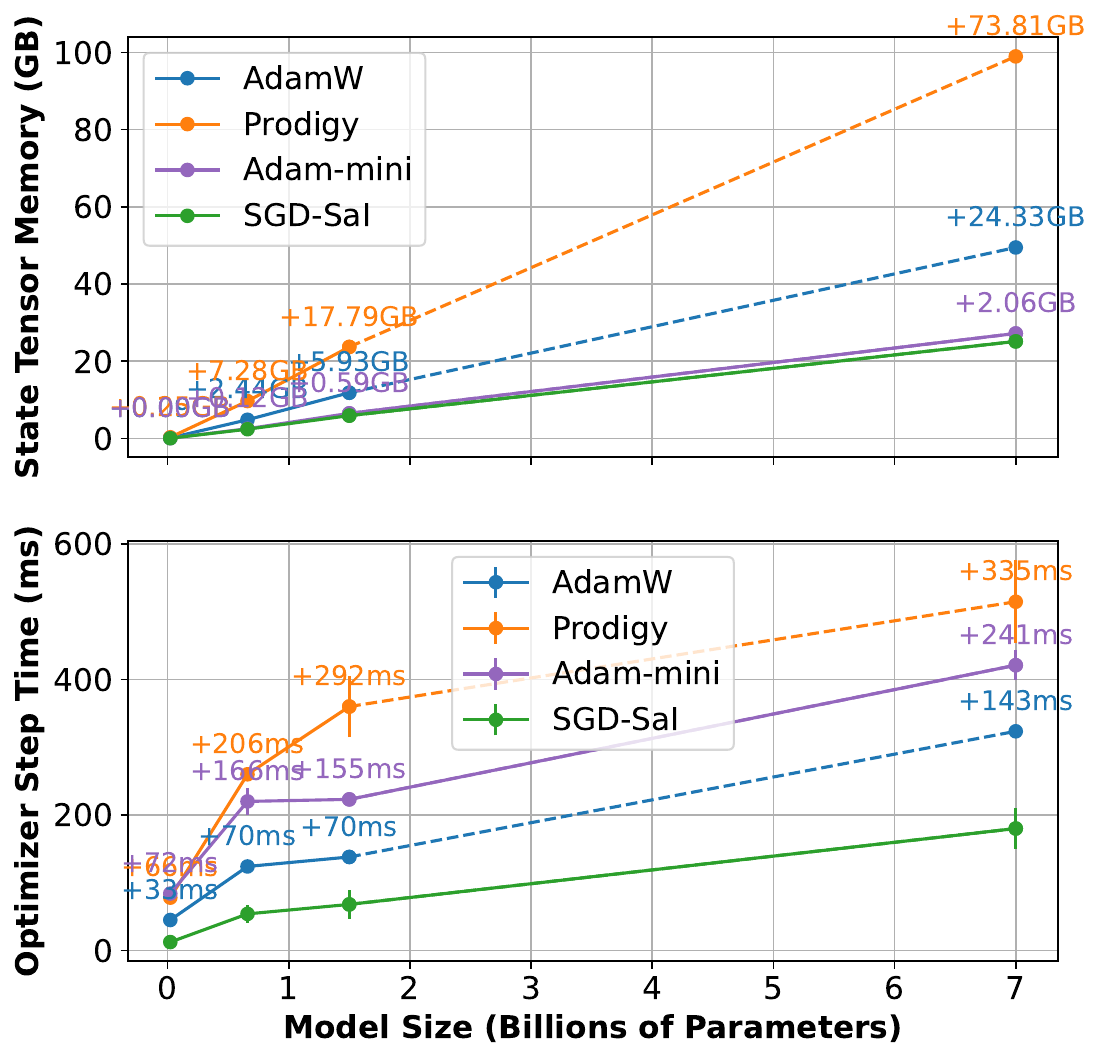}
    \caption{The chart illustrates how memory usage and optimizer step time (in wall-clock time) increase with larger model sizes. It highlights the substantial memory overhead of storing optimizer states as model sizes grow. SGD-SaI exhibits significantly lower memory usage than AdamW and has the shortest optimization step runtime. This runtime refers to the wall clock time required for the optimizer step function. All statistics were measured on a single NVIDIA A100-80GB.}
    \label{fig:state_tensor_memory}
\end{figure}

\begin{figure*}
    \centering
    \includegraphics[width=1.\linewidth]{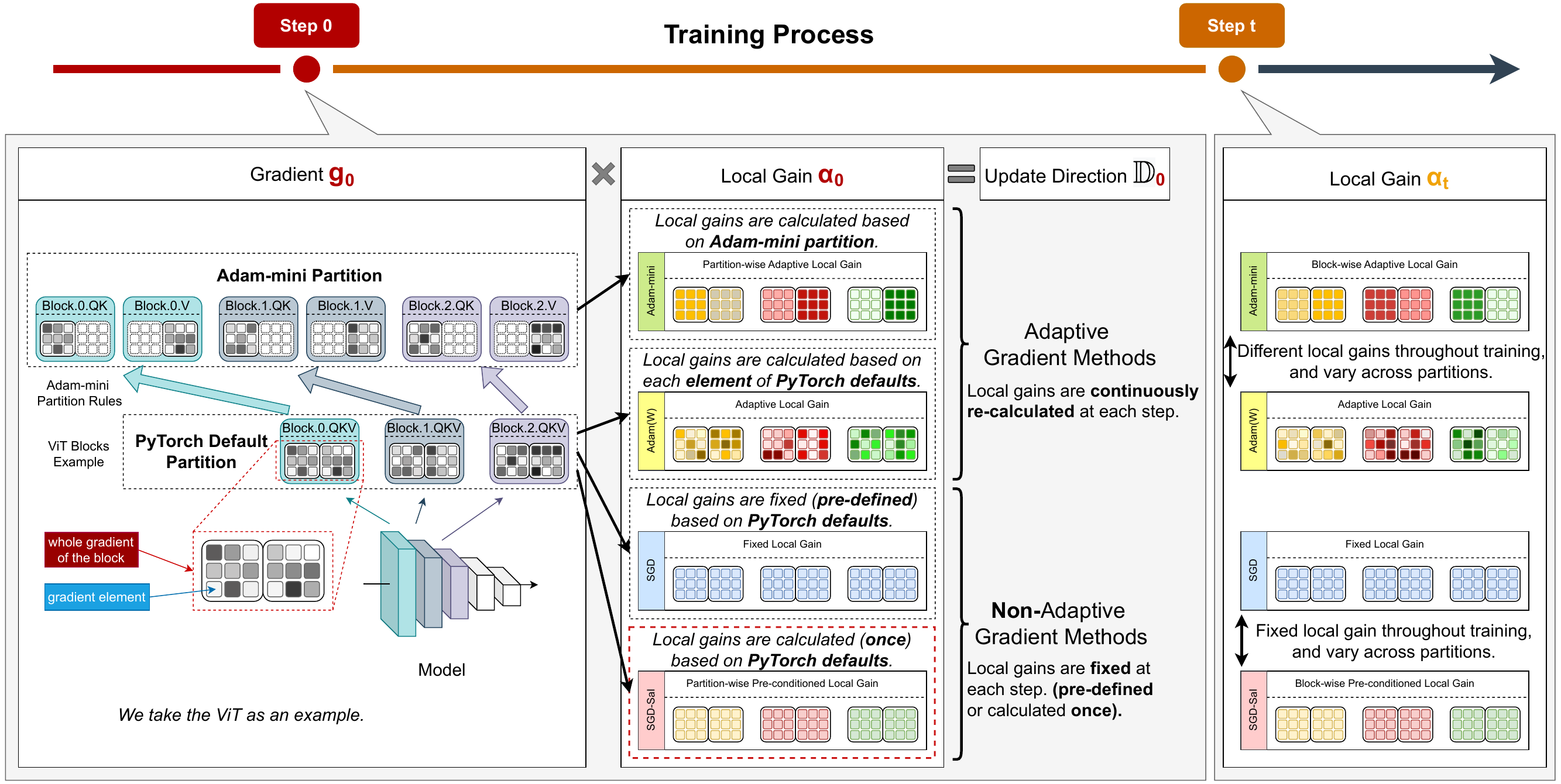}
    \caption{This graph illustrates the differences in local gain behaviours exhibited by four optimizers throughout the training process. We present two popular adaptive gradient methods: Adam(W) and the memory-efficient Adam-mini. The local gains for these methods are recalculated continuously at each step based on the gradients. In contrast, SGD and SGD-SaI are both non-adaptive methods, meaning their local gains remain fixed throughout the training.}
    \label{fig:overview}
\end{figure*}

\section{Introduction}
\label{sec:intro}

Stochastic gradient-based optimization methods, such as Stochastic Gradient Descent (SGD), are fundamental to modern machine learning, enabling the successful training of models across a wide range of scientific and engineering applications. However, training objectives and data are often noisy in practice, and gradients may become sparse due to the inherent characteristics of regularization or specific architectural designs. Moreover, architectural differences can introduce imbalances in the learning dynamics across different parameters. To address these challenges, adaptive gradient methods~\cite{cohen2024adaptivegradientmethodsedge} have been developed to handle better non-stationary objectives, noisy data, and sparse gradients. Among these methods, Adam~\cite{kingma2014adam} and AdamW~\cite{loshchilov2019decoupled} have become indispensable for training Transformer-based models, including Large Language Models (LLMs)~\cite{radford2019language, team2023gemini} and Diffusion Models (DMs)~\cite{ho2020denoising, rombach2022highresolutionimagesynthesislatent}. Their popularity stems from their relative robustness and efficiency in optimizing high-dimensional parameter spaces. The core mechanism of Adam's adaptability lies in its second-order momentum term, $v$,  which acts as a local gain~\cite{hinton2012neural}, dynamically adjusting the learning rate for each parameter. This mechanism enables Adam to perform effectively even in noisy or sparse gradients, addressing imbalances in the learning process across different parameters.

However, this adaptability comes with significant costs when the model size scales up. Specifically, Adam requires storing and updating each parameter's first-order (mean) and second-order (variance) momentum terms. This increases memory usage by at least 3x compared to the parameter size alone. For instance, training a 7-billion parameter model in FP32 using Adam requires approximately 50 GB of memory for the state tensors, a significant challenge even with high-end hardware like NVIDIA A100-80G GPUs. Compared to SGD, the memory demand of Adam can be at least double~\cite{zhang2024adamminiusefewerlearning}, posing a severe limitation on the scalability of deep learning research.

Numerous previous works have sought to reduce memory usage by simplifying optimizer states while preserving the adaptive gradient term to address the memory bottleneck while maintaining the effectiveness of adaptive methods. Approaches such as 8-bit Adam~\cite{DBLP:journals/corr/abs-2110-02861}, Adafactor~\cite{shazeer2018adafactor}, and sign-based methods~\cite{bernstein2018signsgdcompressedoptimisationnonconvex, kunstner2023noise} focus on quantizing or sparsifying the optimizer states. Meanwhile, Adam-mini~\cite{zhang2024adamminiusefewerlearning} introduces parameter block grouping to share adaptive learning rates, leveraging Hessian structure insights~\cite{zhang2024transformers} to reduce memory usage. However, these methods often come with trade-offs. Many risk a performance downgrade compared to AdamW. From an efficiency standpoint, these approaches also introduce additional update complexity. Simplified state tensors still require computations based on full gradients for each parameter at each time step, increasing the overall computational burden. Adam-mini, in particular, necessitates fine-grained parameter partitioning~\cite{zhang2024adamminiusefewerlearning}, further complicating its implementation. As a result, these limitations lead to longer optimizer step times, ultimately slowing down the training process.

In this work, we challenge the necessity of adaptive gradient methods for model training and propose a memory- and computation-efficient alternative. We begin by revisiting the foundational motivation behind Adam's use of second-order momentum. Inspired by the concept of the gradient \textbf{Signal-to-Noise Ratio (g-SNR)}~\cite{xiang2023exploiting}, which quantifies the relationship between a gradient's norm and variance, we leverage this metric to analyze and measure gradient distribution differences across parameters. Through empirical analysis, we investigate the temporal consistency for g-SNR during training and explain why this value can be determined at first training iterations. Furthermore, we analysed the g-SNR distribution across different ViT parameters and explored the g-SNR value correlated with varying parameters of type and its architecture characteristics. Building on this, we argue that g-SNR can be leveraged to adjust learning rate scales, balancing the learning progress based on the distribution of gradients. Incorporating a pre-conditioned learning rate scale computed during the first training iteration, called Scaled at Initialization(SaI), facilitates stable training progress without incurring the memory and computational overhead associated with adaptive gradient terms. We call our method \textbf{SGD-SaI}, a novel optimization approach that eliminates the need for adaptive gradient methods, treating them as simple yet effective updates compared to SGD. In summary, our contributions are as follows:

\begin{itemize}
    \item We challenge the necessity of adaptive gradient methods, specifically identified the existing challenges on Adam-like methods and proposed to use constant g-SNR value to replace the second-order momentum to reduce both the memory and computation cost, called \textbf{Scaled at Initialization(SaI)}.
    \item We empirically analysed the statistics of g-SNR on parameters during training and identified its characteristics over time and distribution over parameters. 
    \item We formula our insight into proposed methods, SGD-SaI, solved the long-stand challenge that SGD can not successfully train tasks with transformer architectures and observed outstanding performance in ViT and decoder-only transformer (LLMs).
    \item We extend our empirical analysis to other popular and practical task training, such as LoRA training on LLMs and Diffusion Models(DMs) and traditional CNN tasks. We observed consistent improvement compared to existing SOTA optimizers.
\end{itemize}

\section{Related Work}
\label{sec:related_work}

\textbf{Adaptive Gradient Methods:}Stochastic gradient descent (SGD) is an efficient optimization method commonly used in deep learning, but it \textbf{struggles with tasks that have non-stationary objectives or involve very noisy and/or sparse gradients}~\cite{kingma2014adam}, often requiring extensive hyperparameter tuning.
To improve upon these limitations, adaptive gradient methods were developed to continuously and dynamically adjust learning rates for individual parameters throughout the training process~\cite{JMLR:v12:duchi11a, graves2014generatingsequencesrecurrentneural, zeiler2012adadeltaadaptivelearningrate}, with the Adam optimizer becoming particularly popular. Adam combines features from AdaGrad~\cite{ward2020adagrad}, which effectively manages sparse gradients, and RMSProp~\cite{hinton2012neural}, which is suitable for online and non-stationary tasks, allowing it to outperform SGD in many cases with less tuning effort.
However, Adam has its own challenges, leading to the creation of enhancements such as AdamW~\cite{loshchilov2019decoupled}, which introduces decoupled weight decay for better generalization, and adaptations~\cite{dozat.2016} that incorporate Nesterov momentum for faster convergence. To address early training noise, warm-up phases and Rectified Adam~\cite{liu2021variance} have been proposed. Additionally, Adaptive Weight Decay~\cite{ghiasi2023improvingrobustnessadaptiveweight} further improves convergence, while ~\cite{mishchenko2023prodigy} introduced a dynamic component for automatic learning rate adjustments within the Adam framework.

\textbf{Adam in Transformer Realm:} Transformers~\cite{vaswani2017attention} have become essential in modern deep learning, particularly in natural language processing. While the Adam~\cite{kingma2014adam} optimizer generally outperforms Stochastic Gradient Descent (SGD) in training Transformer architectures~\cite{xiao2021earlyconvolutionshelptransformers}, it has a significant downside: as model sizes grow, Adam's memory requirements, which are twice that of SGD due to first and second-order momentum storage~\cite{kingma2014adam}, become a concern.
To mitigate this overhead, researchers have explored methods like sign-based optimization~\cite{bernstein2018signsgdcompressedoptimisationnonconvex, kunstner2023noise} and low-precision quantization~\cite{li2023memoryefficientoptimizers4bit, DBLP:journals/corr/abs-2110-02861, dettmers2022llm, dettmers2023case}, although these can compromise performance. Studies have shown that Adam’s adaptive learning rates based on gradient norm history contribute to its performance advantage~\cite{zhang2024transformers}, whereas SGD lacks this capability. However, finding the right learning rate scale for SGD to match Adam’s performance remains unresolved.
Adam's insights, rooted in RMSprop~\cite{hinton2012neural}, suggest that \textbf{a global learning rate should be adjusted according to local gains}. Researchers have developed block-wise dynamic learning rates that perform comparably to Adam with reduced memory use~\cite{zhang2024adamminiusefewerlearning}. Similar trends are seen in parameter-efficient fine-tuning, emphasizing the importance of \textbf{local gains} for learning rate adjustments~\cite{zhang2024riemannian}.
Furthermore, theoretical analyses have raised doubts about the necessity of adaptive gradient methods. While Adam offers practical benefits, research~\cite{liconvergence23} indicates that the convergence rates of Adam and SGD are not significantly different.

\textbf{Gradient at Initialization:} 
Recent research has highlighted the importance of gradient patterns at initialization, demonstrating a strong correlation between these early signals and a model’s eventual performance. Pruning at Initialization (PaI) methods, inspired by the lottery ticket hypothesis~\cite{frankle2018lottery}, leverage this principle by identifying high-potential subnetworks before training begins. These techniques typically remove parameters associated with the lowest gradients or the weakest early learning responses~\cite{tanaka2020pruning, frankle2020pruning, lee2018snip}, emphasizing how initial gradient-based criteria can guide the formation of effective, sparse architectures.

From a gradient sparsity perspective, PaI methods effectively preserve the essential characteristics of the full network’s gradient distribution. The resulting subnetworks maintain similar gradient variance and overall gradient magnitude by masking out parameters tied to minimal gradient or learning response. This careful selection ensures that the pruned models exhibit performance levels on par with their unpruned counterparts despite operating with significantly fewer parameters.

A similar observation has also been revealed in Zero-Cost NAS studies~\cite{abdelfattahzero, lizico, xiang2023exploiting}, which aim to predict the performance of untrained networks by analyzing gradient patterns, finding that gradient score rankings—such as the gradient sum—correlate more strongly with architectural structures than with data batches or initialization parameters. Research by ~\cite{bhardwaj2021doestopologyinfluencegradient} highlights that gradient flow patterns are inherently linked to a network's architecture. Additionally, studies~\cite{lizico, xiang2023exploiting} show that gradient sparsity, measured by mean and variance, is closely related to convergence rates and generalization ability. They emphasize calculating gradient sparsity block-wise due to the diverse distributions of gradients across parameter blocks. Moreover, \cite{lei2023balanceessenceacceleratingsparse} suggests that a balanced training procedure with low-variance gradients enhances sparse training.

\section{Problem Setting}
\textbf{Notations.} 
A neural network is defined based on a set of trainable parameters in specific architectures. We denote the neural network's parameters as $\theta \in \mathbb{R}^d$, where 
$d$ is the total number of parameters. The training loss function $L(\theta)$ defines the objective to be minimized. The parameter space is partitioned into $B$ blocks based on the definition of network architectures, denoted as $\theta^{(i)} \in \mathbb{R}^{d_i}$ for $i \in \{1, 2, \dots, B\}$, where $d = \sum_{i=1}^{B}d_i$. Each parameter $\theta^{(i)}_j$ within block $i$ for $j\in[d_i]$ is associated with its own gradient $g^{(i)}_j = \nabla_{\theta^{(i)}_j} L(\theta)$.

Key notations used throughout the paper are as follows: We denote $t\in\mathbb{N}$ as the index for the training step, $\eta > 0$ as the global learning rate, $\lambda \geq 0$ as the weight decay coefficient, $\mu$ as the momentum coefficient, $g^{(i)}_t \in \mathbb{R}^{d_i}$ as the gradient of the loss w.r.t. $\theta^{(i)}$ at step $t$. $[d_i]$ is the index set $\{1, 2, 3, \dots, d_i\}$ corresponding to the parameters in block $i$. And $\mathcal{O}(*)$ means the complexity, here we use it to measure the storage.

\textbf{Stochastic Gradient-based Optimization:} 
Given the loss function $L(\theta)$, the goal of a general optimization process is to update $\theta$ in the following form iteratively:
\begin{equation}
    \theta_{t+1} = \theta_t - \eta_t \mathbb{D}_t,
\end{equation}
where $\mathbb{D}_t$ denotes the update direction at step $t$. The choice of $\mathbb{D}_t$ defines the specific optimization algorithm. For SGD, the update direction is defined as the negative gradient of the loss with respect to $\theta_t$:
\begin{align}
    \mathbb{D}_t = g_t
    \textit{, where }
    g_t = \nabla L(\theta_t)
\end{align}

The first-order momentum term was introduced to SGD to enhance the optimisation process, and it is called SGD with momentum(SGDM)~\cite{Nes83}. The momentum $m$ can be defined as:
\begin{align}
    m_t = \beta_1 m_{t-1} + (1-\beta_1) g_t
\end{align}
The update becomes:
\begin{align}
    \mathbb{D}_t&= m_t
\end{align}
This addition helps accelerate convergence by incorporating information from previous gradients to smooth out the update steps. Specifically, it reduces oscillations in the optimization trajectory, particularly in scenarios with steep or narrow ravines in the loss landscape. By maintaining a running average of past gradients, the momentum term allows SGD to move more consistently in directions that lead to faster convergence, addressing challenges like slow progress on flat regions of the loss surface.

\textbf{Adaptive Gradient Methods:} Adaptive gradient methods like Adam adopted first-order momentum \(m_t\) as we mentioned above while introducing the second-order momentum \(v_t\), which tracks squared gradients to adjust the learning rate for each parameter, the \(v_t\) defined as:
\begin{align}
    v_t = \beta_2 v_{t-1} + (1-\beta_2) g_t^2.
\end{align}
The update direction for Adam is as follows:
\begin{align}
    \mathbb{D}_t&= \alpha_t m_t,
    \textit{ where }
    \alpha_t = \frac{1}{\sqrt{\hat{v}_t} + \epsilon}.
\end{align}
$\hat{v}_t$ term is the $v_t$ with bias correction.
Notably, $\alpha_t$ is the local learning rate gain (aka. adaptive learning rate). The key computational challenge is the storage and updating of 
$v_t$, which requires $\mathcal{O}(d)$ additional memory.

\textbf{Memory Efficient Adam:}
As Scaling Law ~\cite{kaplan2020scaling} introduced, Transformer model sizes in recent days have significantly increased compared to the model size when Adam was introduced. Consequently, the memory overhead of the Adam optimizer has become a significant issue, as it requires at least 3x times of memory compared to parameter size. 
Several approaches have been proposed to reduce the memory overhead of the second-order momentum $v_t$, including (a) Adafactor~\cite{shazeer2018adafactor} shares the $v$ across dimensions, reducing storage from $\mathcal{O}(d)$ to $\mathcal{O}(\sqrt{d})$. However, Adafactor trades off memory savings for lower update precision. (b) Low-bit optimisers quantize~\cite{DBLP:journals/corr/abs-2110-02861} the storage of $v_t$ to low-precision formats (e.g., 8-bit) to save memory. While effective, quantization introduces additional implementation complexity. (c) Adam-mini~\cite{zhang2024adamminiusefewerlearning} partitions the block and uses the moving average of the estimated $v_t$ for each block, thereby reducing storage from $\mathcal{O}(d)$ to $\mathcal{O}(B)$. However, Adam-mini not only introduces additional computational costs compared to the Adam update process, but its complex partition policy is also incompatible with the default PyTorch partitioning; for example, in default PyTorch partitioning, the attention QKV considered the same group of parameters,  while Adam-mini requires further partitioning them with based on heads or neurons. Furthermore, it adaptively calculates and updates $\alpha_t$ over time, highlighting its intensive computational complexity. While these approaches reduce memory usage, they all retain the second-momentum term $v_t$ with a trade-off to either performance or update speed.

\textbf{Problem Statements.} We aim to eliminate the need for the explicit second-order momentum $v_t$ entirely while maintaining effective learning rate adaptation. Instead of using $\alpha^{(i)}_t = \frac{\eta}{\sqrt{\hat{v}^{(i)}_t} + \epsilon}$, we aim to design a new rescaling factor $\alpha^{(i)}_t$ that adapts to the loss landscape without requiring the computation or storage of $v^{(i)}_t$.

Given a block-wise parameter $\theta^{(i)}$, we seek a new function $\mathcal{F}$ such that:
\begin{equation}
    \alpha^{(i)}_t = \mathcal{F}(g^{(i)}_1, g^{(i)}_2, \dots, g^{(i)}_t).
\end{equation}
where $\mathcal{F}$ determines the local learning rate gain for each parameter block using the gradient history $[g^{(i)}_1, g^{(i)}_2, \dots, g^{(i)}_t]$.

\section{Methods}

\begin{figure*}[!ht]
    \centering
    \includegraphics[width=0.33\linewidth]{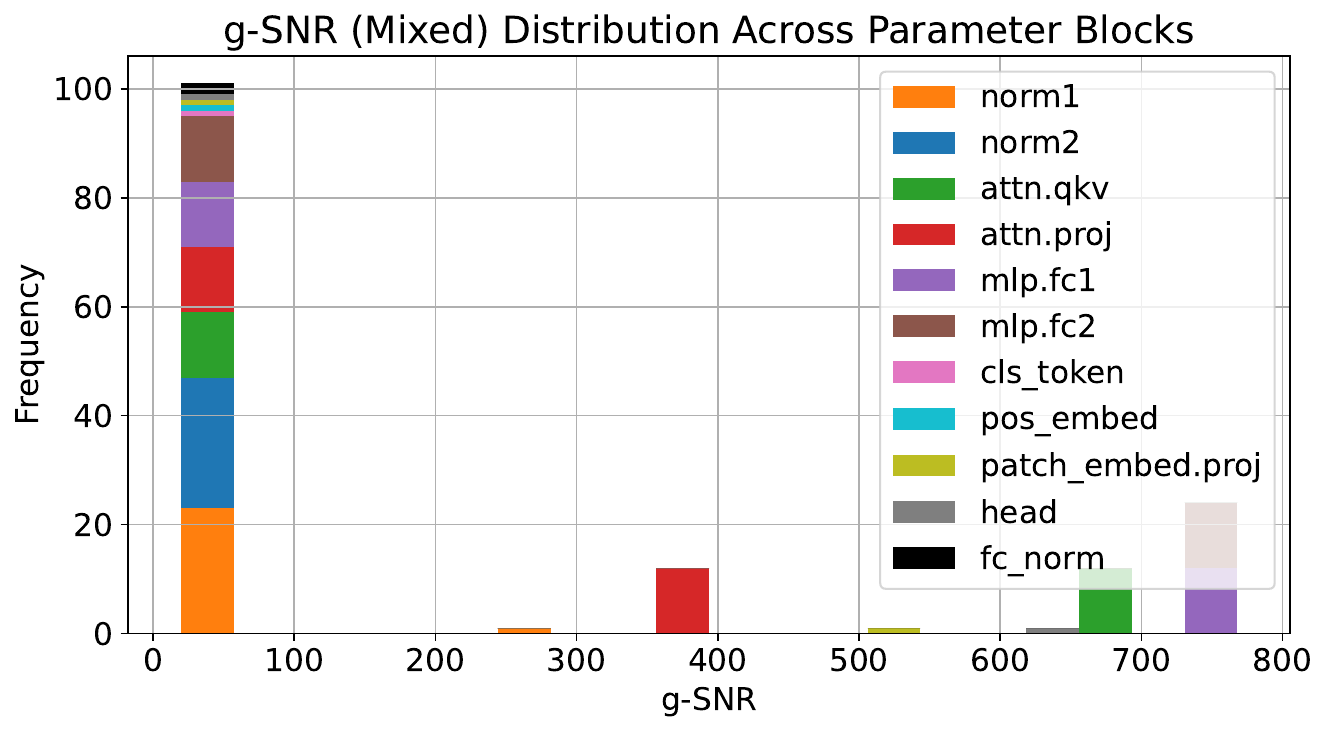}
    \includegraphics[width=0.33\linewidth]{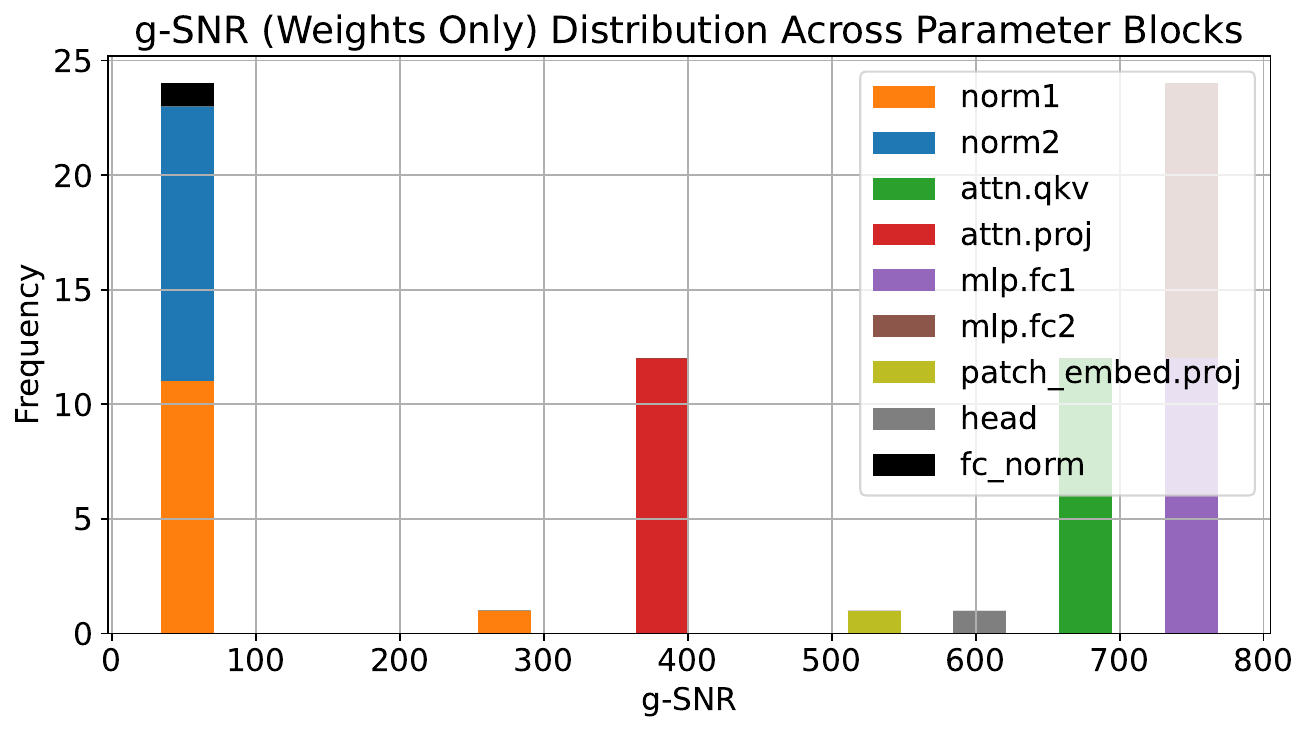}
    \includegraphics[width=0.33\linewidth]{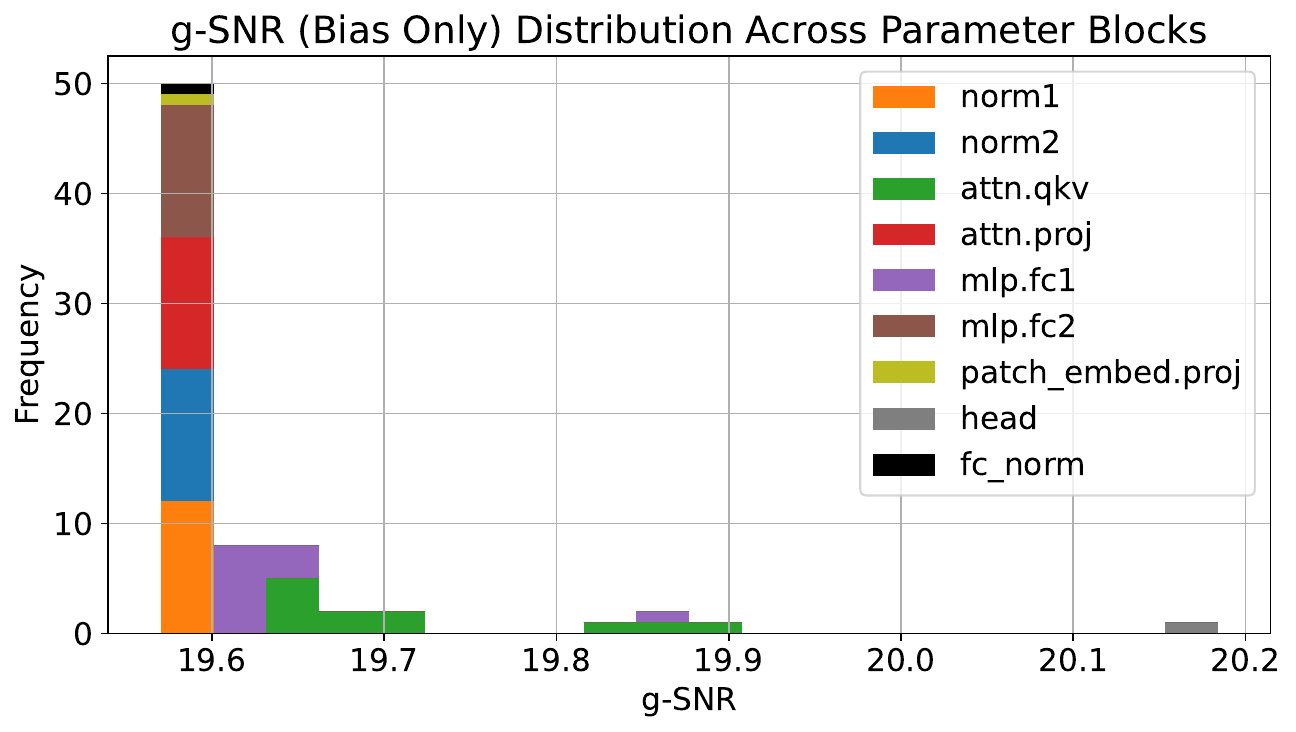}
    \caption{We observe that the g-SNR varies across different parameter blocks. However, for most weights, the parameter blocks that share the same structure across different transformer layers (blocks) tend to have similar g-SNR values. Additionally, the g-SNR values for the bias parameters are consistently low magnitude. Our method can be viewed as partitioning all parameter blocks based on their structure.}
    \label{fig:gsnr_across_blocks_hist}
\end{figure*}

\begin{figure}[!h]
    \centering
    \includegraphics[width=\linewidth]{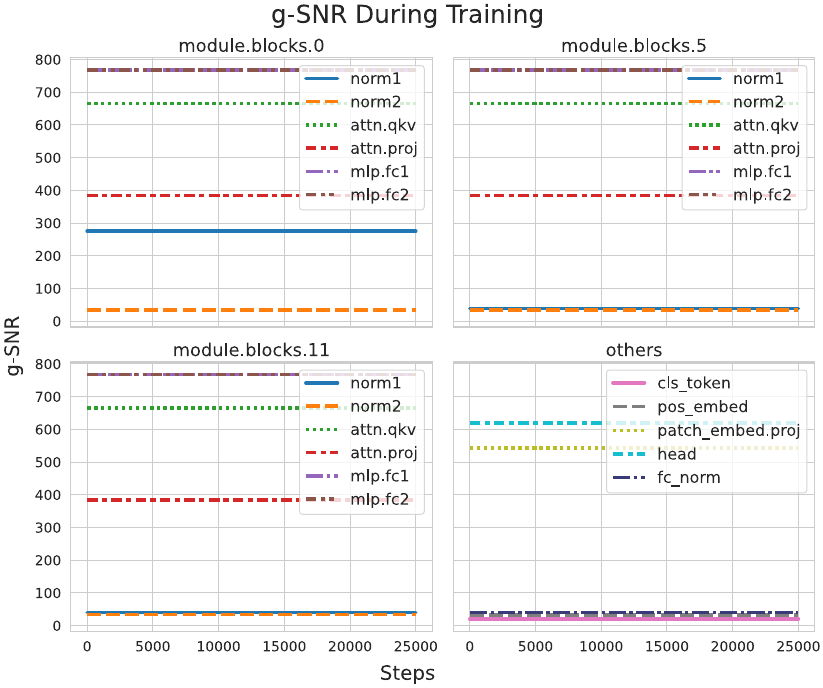}
    \caption{We plot the g-SNR distribution over time for three different transformer blocks: shallow (block 0), middle (block 5), and deep (block 11). Additionally, we analyze some distinct types of parameter blocks. Our observations indicate that while the g-SNR values vary across different parameter blocks, they tend to remain relatively constant over time.}
    \label{fig:gsnr_across_time_plot}
\end{figure}

\begin{figure*}[h!]
    \centering
        \includegraphics[width=.99\textwidth]{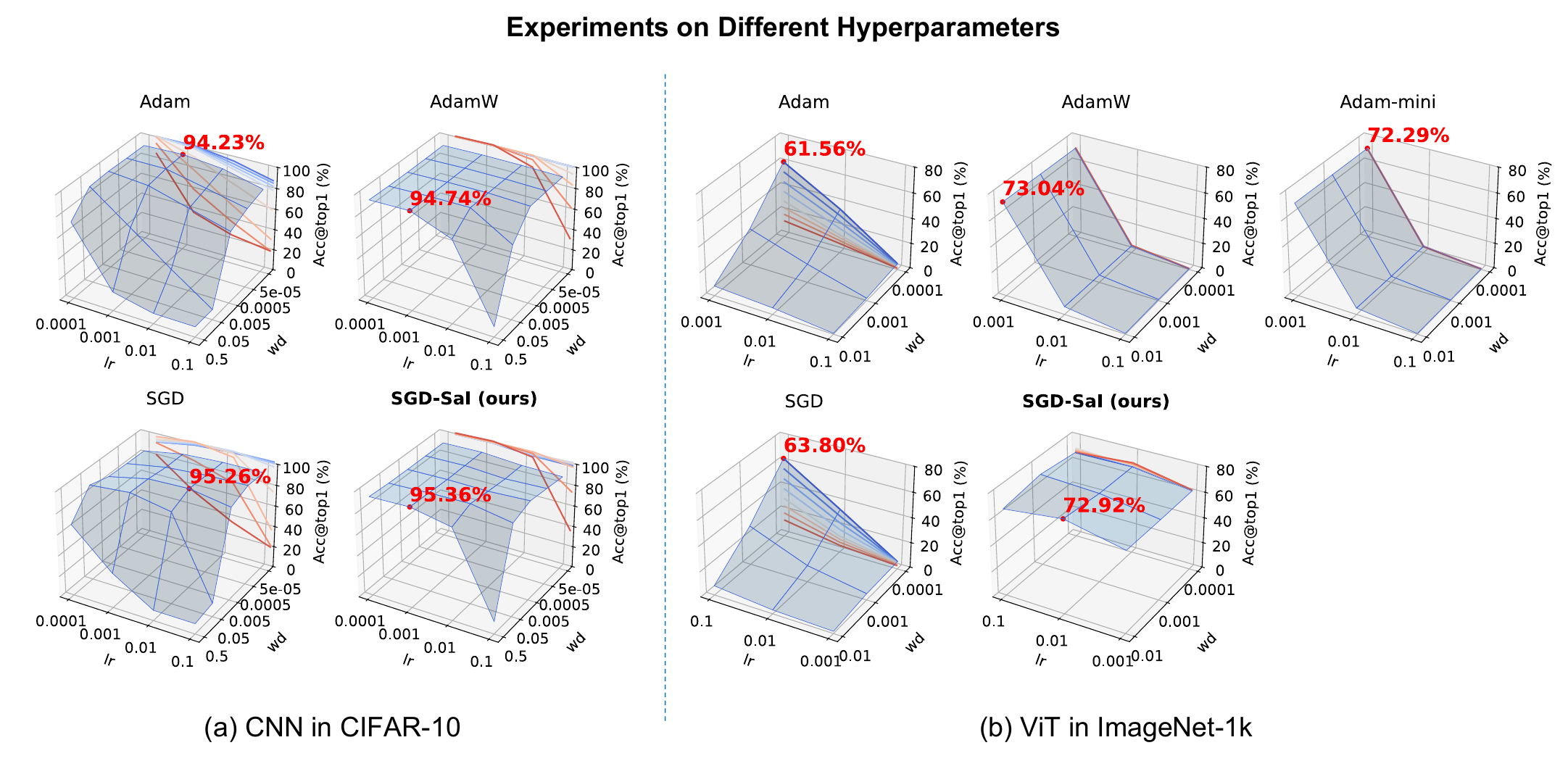}
    \caption{Comparison of top-1 test accuracy distributions for CNNs on CIFAR-10 (Left) and ViTs on ImageNet-1k (Right) across different hyperparameter combinations. Each method demonstrates distinct performance trends, including Adam, AdamW, SGD, and SGD-SaI. Adam-Mini is only compared in the ViT case as its modification target on transformer training. SGD-SaI consistently shows enhanced robustness and performance under varying hyperparameter settings.}
    \label{fig:fig_vit_cnn_performance_3d_surface}
\end{figure*}

Considering the substantial memory overhead introduced by the second-order momentum in the Adam optimizer, this section explores strategies to reduce this cost by revisiting the foundational motivations for adaptive gradient methods.

In the following subsections, we design a memory-efficient learning rate local gain, termed g-SNR, to replace the second-order momentum. We analyse the distribution of g-SNR across different parameter groups throughout the network. This aligns with the motivation of parallel works \cite{zhang2024adamminiusefewerlearning} that focus on partitioning parameter groups for learning rate adjustment. Furthermore, we investigate the behaviour of g-SNR during training, demonstrating how dynamic local gains can be replaced with constant preconditioned values calculated in the initial iterations.

Finally, we introduce our proposed method, SGD-SaI, detailing its design and implementation. This method builds on the insights derived from g-SNR analysis, offering a memory-efficient alternative to second-order momentum while maintaining competitive performance.

\subsection{Memory Efficient Local Gain: g-SNR}
Adam builds upon RMSprop, designed to find a \textbf{local gain} for the learning rate, enabling parameter-specific adjustments within deep neural networks \cite{hinton2012neural, kingma2014adam}. By incorporating second-order momentum, Adam improves upon SGD by better handling problems with non-stationary objectives and tasks characterized by noise or sparse gradients \cite{kingma2014adam}. This mechanism allows Adam to dynamically rescale gradients, effectively adjusting the learning pace across parameter blocks with distinct gradient patterns. Consequently, Adam outperforms SGD when training architectures with heterogeneity problems in the Hessian matrix, such as Transformers \cite{zhang2024transformers, zhang2024adamminiusefewerlearning}. Another key insight arises from the warm-up mechanism: even with second-order momentum, Adam still requires a warm-up phase to reduce the learning rate at the beginning of training, aiming to mitigate gradient variance~\cite{liu2021variance}. During this phase, gradients are known to be sparse and noisy. Reducing the learning rate directly during the warm-up phase effectively lowers gradient variance, stabilizing the training process straightforwardly and efficiently.

Intuitively, adaptive gradient methods dynamically adjust the learning rate for each parameter during training. This mechanism encourages parameters with less learning history to learn more while slowing down the learning pace for parameters progressing too quickly. Essentially, it acts as a compensatory approach to address learning imbalances across parameters after they arise. However, if we can predict and pre-empt these imbalances before they occur, we could potentially eliminate the need for second-order momentum, which relies on learning history to evaluate and correct them. 

Considering the root cause of why learning imbalance occurred across different parameters, we discussed them in two main parts. Firstly, as inherited from the architecture characteristics, the parameter in different layers or with different architectures will receive distinct gradient pattern~\cite{tanaka2020pruning, lizico, xiang2023exploiting}, thus bringing the optimal learning rate for different parameters are distinct and need to be re-adjust with local gain~\cite{hinton2012neural}, Secondly, within the parameter groups, the gradient can be noisy or sparse based on the objective and data, that will introduce imbalance update to parameters.

We propose using the gradient signal-to-noise ratio (g-SNR) introduced by \cite{xiang2023exploiting} to adjust the learning rate block-wisely, as it measures the norm and variance of gradients of the parameter block, which reflects overall update magnitude and variance of gradient between paramters. Specifically for each block~$i$, the gradient norm ($\ell^2$-norm) and variance are calculated as
\[
G_{\text{norm}}^{(i)} = \sqrt{\sum_{j=1}^{d_i} \left( g^{(i)}_j \right)^2}, \quad G_{\text{var}}^{(i)} = \frac{1}{d_i} \sum_{j=1}^{d_i} \left( g^{(i)}_j - \bar{g}^{(i)} \right)^2,
\]
where $\bar{g}^{(i)} = \frac{1}{d_i} \sum_{j=1}^{d_i} g^{(i)}_j$, and $d_i$ is the number of parameters in block~$i$. The gradient signal-to-noise ratio for each block is then given by
\[
G_{\text{snr}}^{(i)} = \frac{G_{\text{norm}}^{(i)}}{\sqrt{G_{\text{var}}^{(i)}} + \epsilon},
\]
where $\epsilon$ is a small constant added for numerical stability. To ensure consistent scaling across all blocks, we normalize the g-SNR of each block by the maximum g-SNR among all blocks:
\[
\tilde{G}_{\text{snr}}^{(i)} = \frac{G_{\text{snr}}^{(i)}}{\max_{k} G_{\text{snr}}^{(k)}}.
\]
This normalization confines the g-SNR values between 0 and 1, facilitating a fair comparison and adjustment of learning rates across different parameter blocks. Thus, we establish the local gain by replacing $v_t$ with the following expression:
\begin{align*}
    \alpha^{(i)}_t = \mathcal{F}(g^{(i)}_t) = \tilde{G}_{\text{snr}}^{(i)}
\end{align*}
where $\alpha^{(i)}_t$ represents the local gain at step $t$ guided by the temporal value $\tilde{G}_{\text{snr}}^{(i)}$  which determines the update direction $\mathbb{D}_t$. This approach reduces the memory overhead of $v_t$ from $\mathcal{O}(d)$ to $\mathcal{O}(B)$. 

Adapting the learning rate according to the normalized gradient signal-to-noise ratio significantly influences gradient variance during training. When the high gradient noise or sparsity in block~$i$ occurs, $\tilde{G}_{\text{snr}}^{(i)}$ tend to have relatively lower value, the learning rate $\eta$ is scaled down by a factor $\alpha^{(i)} = \tilde{G}_{\text{snr}}^{(i)}$, resulting in a reduced learning rate $\eta^{(i)} = \alpha^{(i)} \eta$. This adjustment decreases the magnitude of parameter updates for that block:
\[
\theta^{(i)}_{t+1} = \theta^{(i)}_t - \eta^{(i)} \nabla_{\theta^{(i)}} L(\theta_t).
\]
Lowering the learning rate mitigates the amplification of gradient noise, thereby reducing gradient variance within each training step, leading to smoother convergence and enhanced robustness \cite{liu2019variance}. If $\tilde{G}_{\text{snr}}^{(i)}$ remains low across multiple batches, the continued reduction of $\eta^{(i)}$ further stabilizes training by preventing large, erratic updates.

\subsection{Statistics Analysis for g-SNR}
Building on the insights above, we implemented the g-SNR mechanism using PyTorch's Default Partition \cite{zhang2024adamminiusefewerlearning}, which computes g-SNR within each parameter block and dynamically re-scales the learning rate accordingly. To assess its effectiveness, we conducted experiments on Vision Transformer (ViT) pre-training tasks using ImageNet-1K, selecting ViT/S-16 for comprehensive tracing and analysis of gradient patterns throughout the training process.

Our analysis revealed that g-SNR remains relatively stable over time while exhibiting distinct patterns across different parameter classes, as shown in Fig.~\ref{fig:gsnr_across_time_plot}. Specifically, we examined transformer blocks from shallow, middle, and deep layers within the network and parameters outside the transformer blocks, such as positional embeddings. 

Given the g-SNR definition we provide in the previous subsection, we analyze its behaviour as follows: As modern initialization schemes (e.g., Xavier~\cite{kumar2017weight}, Kaiming~\cite{he2015delving}) ensure that at \( t=0 \):
\[
G_{\text{norm}}^{(i)}(0) \quad \text{and} \quad G_{\text{var}}^{(i)}(0)
\]
are well-controlled. This implies that \(G_{\text{snr}}^{(i)}(0)\) starts from a stable, architecture-driven ratio. During the training process, parameters are updated and controlled by the step size \(\eta\) is the learning rate. Assuming \(\eta\) is sufficiently small to stabilize training process, we have \( \theta^{(i)}_{t+1} \approx \theta^{(i)}_{t} \). Thus, the change in parameters per iteration is small.Consider the gradient at iteration \( t+1 \):
\[
\mathbf{g}^{(i)}_{t+1} = \nabla_{\theta^{(i)}}L(\theta_{t+1}).
\]
Consider a first-order Taylor expansion of the gradient around \(\theta^{(i)}(t)\):
\[
\mathbf{g}^{(i)}_{t+1} \approx \mathbf{g}^{(i)}_{t} + J^{(i)}_{t}\Delta \theta^{(i)}_{t},
\]
where \(J^{(i)}_{t}\) is the Jacobian 
(or a first-order sensitivity matrix) of \(\mathbf{g}^{(i)}\) w.r.t. \(\theta^{(i)}\), and \(\Delta \theta^{(i)}_{t}=\theta^{(i)}_{t+1}-\theta^{(i)}_{t}\). Since \(\|\Delta \theta^{(i)}_{t}\|\) is small, the change in the gradient vector is also small. Hence,
\[
g_{j(t+1)}^{(i)} \approx g_{j(t)}^{(i)}, \quad \forall j.
\]
Because each component \( g_{j(t+1)}^{(i)} \) differs only slightly from \( g_{j(t)}^{(i)} \), their average and variance remain stable:
\[
\bar{g}^{(i)}_{t+1} \approx \bar{g}^{(i)}_{t}, \quad
G_{\text{var}(t+1)}^{(i)} \approx G_{\text{var}(t)}^{(i)}.
\]

Similarly, for gradient norm,
\[
G_{\text{norm}(t+1)}^{(i)} = \sqrt{\sum_{j=1}^{d_i} \bigl(g_{j(t+1)}^{(i)}\bigr)^2} \approx G_{\text{norm}(t)}^{(i)}.
\]

Since both \( G_{\text{norm}(t)}^{(i)} \) and \( G_{\text{var}(t)}^{(i)} \) remain nearly unchanged,
\[
G_{\text{snr}(t+1)}^{(i)} = \frac{G_{\text{norm}(t+1)}^{(i)}}{\sqrt{G_{\text{var}(t+1)}^{(i)}}+\epsilon} \approx \frac{G_{\text{norm}(t)}^{(i)}}{\sqrt{G_{\text{var}(t)}^{(i)}}+\epsilon} = G_{\text{snr}(t)}^{(i)}.
\]

Thus, \(G_{\text{snr}(t)}^{}\) remains effectively constant over iterations. Even though parameters change, the "shape" or statistical profile of the gradient distribution does not drastically alter. The g-SNR measures a dimensionless ratio that characterizes this shape. Minor parameter shifts do not significantly affect this ratio; hence, it remains nearly constant.

This finding aligns with the observation by \cite{xiang2023exploiting} that g-SNR strongly correlates with architecture. Leveraging this insight, we replaced the dynamic calculation of g-SNR with constant values determined during initialization, significantly reducing computational costs during each training step.

When calculating g-SNR using PyTorch's Default Partition, we observed that the g-SNR values vary significantly across partitions. By leveraging constant g-SNR values, this approach effectively assigns a pre-conditioned learning rate scale to each partition. \cite{zhang2024adamminiusefewerlearning} highlights a key limitation of PyTorch's default parameter partitioning: its lack of granularity for optimizers like Adam-mini. While PyTorch groups parameters such as attention QKV together, Adam-mini requires finer partitions, such as by attention heads or neurons, to perform effectively, especially in Transformer-based architectures. This limitation stems from the default partitioning's failure to align with Hessian sub-block structures critical for optimization.

This observation does not hold true in our case. Our empirical results, shown in Fig.~\ref{fig:fig_vit_cnn_performance_3d_surface} and discussed further in Sec.~\ref{sec:results}, demonstrate that our method works effectively with PyTorch's Default Partition and does not require any additional fine-grained partitioning strategies. The distribution of g-SNR across different partitions is detailed in Fig.~\ref{fig:gsnr_across_blocks_hist}, where we observe that, for most weights, parameter blocks sharing the same structure across different Transformer layers exhibit similar g-SNR values. Additionally, the g-SNR values for bias parameters remain consistently low, reflecting their uniform magnitude. A notable exception is the $norm1$ weights from $blocks.0$, which connect to the input from embedded patches, whereas all other $norm1$ weights connect to the output of the previous block. This observation highlights that our g-SNR values can effectively identify distinct characteristics among different parameter groups and the network's topological impacts. Moreover, it indicates that gradient sparsity and noise levels vary across parameter groups, confirming the necessity of using a local gain mechanism to balance learning rates across partitions.

Notably, our approach, compatible with PyTorch's Default Partition, enables simultaneous updates of each coarse-grained parameter block and eliminates the need for dynamic learning rate calculations. This efficiency resulted in a threefold speedup in the optimizer update step compared to Adam-mini when training the GPT2-small model. Moreover, it reduces the implementation complexity associated with the exhaustive Hessian calculations required for fine-grained parameter partitioning \cite{zhang2024adamminiusefewerlearning}.

In summary, instead of relying on second-order momentum to compute gradient history and adjust learning rates to address imbalanced updates after they occur, our g-SNR approach determines the gradient sparsity level at the first iteration of training. This enables assigning appropriate pre-conditioned learning rate scales to different parameter partitions, simplifying the update process, improving memory efficiency, and significantly speeding up optimization.

\subsection{Proposed Methods Detail: SGD-SaI}
We propose a new method called \textbf{SGD-SaI} that removes adaptive gradient components by rescaling the learning rates of each parameter block using the g-SNR calculated from the initial batch. The algorithm details are presented in Algorithm~\ref{algo:algorithm_cmp}. By leveraging the initial g-SNR, we capture the inherent gradient characteristics of different parameter blocks, allowing for a constant scaling factor that addresses the variations in gradient magnitudes across blocks.

As our method eliminates the dynamic terms associated with adaptive gradient algorithms, it only introduces a few computations at the first iteration compared to naive Stochastic Gradient Descent with Momentum (SGDM). Specifically, the additional computation involves calculating the g-SNR for each parameter block during the initial batch. After this initial computation, the training proceeds similarly to standard SGDM, making our method computationally efficient and comparable in complexity to traditional SGD.

To update the g-SNR based on the actual gradient sparsity without affecting the gradient computation, we adopt \textbf{decoupled weight decay} as proposed by Loshchilov and Hutter~\cite{loshchilov2019decoupled}. Decoupled weight decay applies regularization directly to the parameters rather than incorporating it into the gradient computation. This approach is equivalent to regularization in SGD and allows us to accurately compute the gradient statistics needed for the g-SNR without the weight decay term distorting the gradient values. By doing so, we ensure that the g-SNR reflects the gradients' true sparsity and noise characteristics.

Our implementation remains extremely straightforward, as we adopt the simplest approach that requires only minimal modifications to the existing SGD optimizer. This simplicity ensures that existing tricks and frameworks that support SGD can seamlessly integrate with and support our method.
\begin{figure}[!ht]
    \centering
    \begin{minipage}{0.45\textwidth}
        \begin{algorithm}[H]
            \caption{SGD-SaI}\label{algorithm:sgd_boost}
            \footnotesize
            \begin{algorithmic}[1]
                \Require $T$ (total steps), $\eta$ (learning rate), $\theta^i$ ($i$-th parameter block), $L(\theta)$ (loss function), $\lambda$ (weight decay), $\mu$ (momentum), $\epsilon$ (small constant), $maximize$
                \For{$t \gets 1$ to $T$}
                    \State Compute gradient: $g^i_t \gets \nabla_{\theta^i} L(\theta_{t-1})$
                    \If{$maximize$}
                        \State $g^i_t \gets -g^i_t$
                    \EndIf
                    \State \text{/* Apply momentum */}
                    \If{$t > 1$}
                        \State $m^i_t \gets \mu m^i_{t-1} + (1 - \mu) g^i_t$
                    \Else
                        \State $m^i_t \gets g^i_t$
                        \State \text{/* Compute g-SNR */}
                        \State $G_{\text{snr}}^i \gets \dfrac{G_{\text{norm}}^i}{\sqrt{G_{\text{var}}^i} + \epsilon}$
                        \State \text{/* Normalize g-SNR */}
                        \State $\tilde{G}_{\text{snr}}^i \gets \dfrac{G_{\text{snr}}^i}{\max_k G_{\text{snr}}^k}$
                    \EndIf

                    \State \text{/* Apply weight decay */}
                    \State $\theta^i_t \gets \theta^i_{t-1} - \lambda \eta \theta^i_{t-1}$
                    \State \text{/* Update parameters with scaled learning rate */}
                    \State $\theta^i_t \gets \theta^i_t - \eta \tilde{G}_{\text{snr}}^i m^i_t$
                \EndFor
            \end{algorithmic}
        \end{algorithm}
    \end{minipage}
    \caption{Our Algorithm. we introduce a simple parameter-block-wise scaling using the normalized g-SNR to rescale the learning step size. This allows SGD to perform block-wise effective learning, unlocking its potential to work well on networks with block heterogeneity problems~\cite{zhang2024transformers}.}
    \label{algo:algorithm_cmp}
\end{figure}

\begin{figure*}[!ht]
    \centering
    \includegraphics[width=0.33\linewidth]{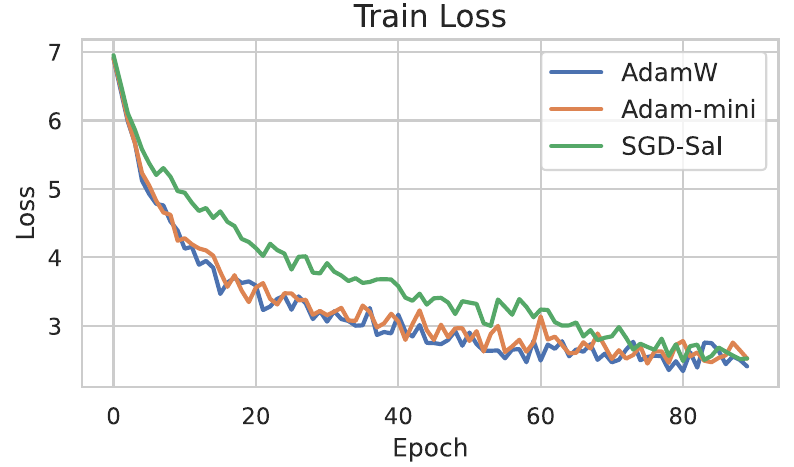}
    \includegraphics[width=0.33\linewidth]{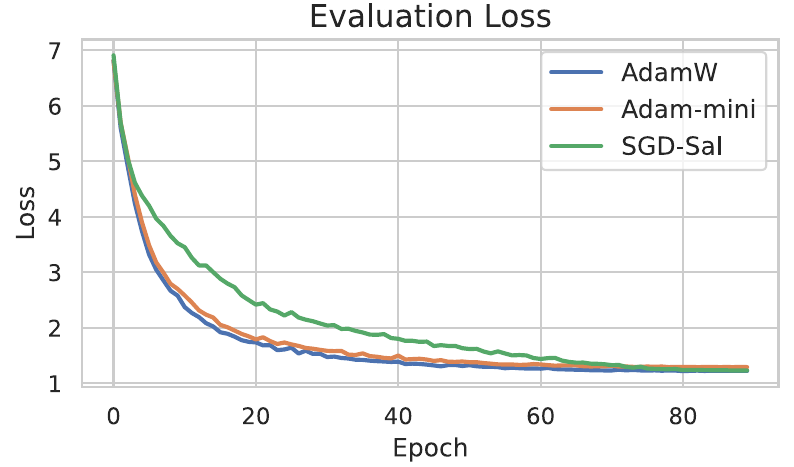}
    \includegraphics[width=0.33\linewidth]{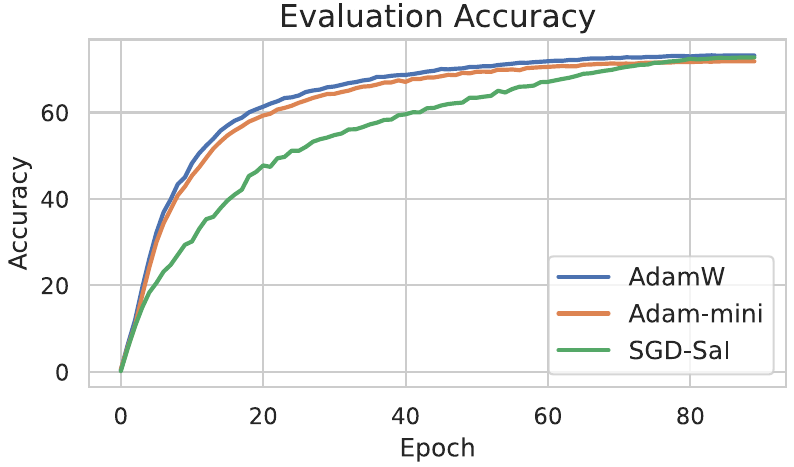}
    \caption{This figure displays the training and evaluation loss and accuracy of the ViT on ImageNet1k). Although our method has a slower convergence speed, we can still achieve comparable performance by the end of the training process. Additionally, our approach is designed to have a lower memory footprint and a faster optimization speed. }
    \label{fig:vit_loss_acc_across_time}
\end{figure*}

\begin{figure}[!ht]
    \centering
    \includegraphics[width=\linewidth]{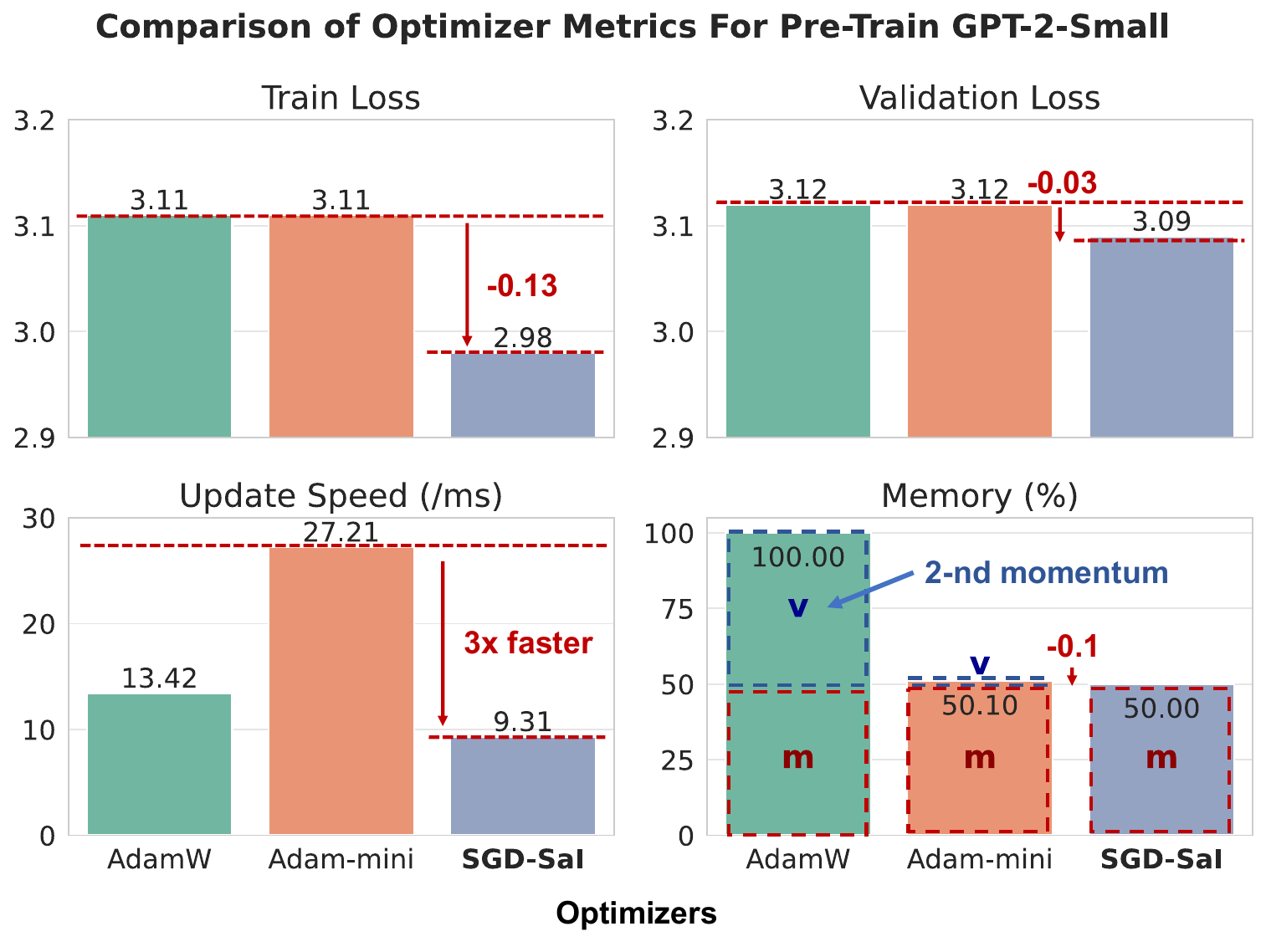}
    \caption{Metrics comparison of optimizers (AdamW, Adam-mini, and SGD-SaI) during pre-training of GPT-2 Small. The figure includes four subplots: (a) Train Loss shows that SGD-SaI achieves the lowest train loss, outperforming Adam-mini by 0.13. (b) Validation Loss illustrates a slight improvement in SGD-SaI with a reduction of 0.03 compared to Adam-mini. (c) Update Speed highlights that SGD-SaI is three times faster than Adam-mini, with AdamW showing moderate performance. (d) Memory Usage indicates that AdamW consumes 100\% memory, while both Adam-mini and SGD-SaI utilize approximately half, demonstrating better efficiency. Annotated values provide clarity on performance metrics, with red highlights emphasizing improvements from Adam-mini.}
    \label{fig:optimizer_metrics_comparison}
\end{figure}

\section{Experiments}
\label{sec:results}

This section evaluates our method through several tasks, including pre-training for Large Language Model (LLM) and Vision Transformer (ViT), Parameter-Efficient Fine-Tuning (PEFT) tasks on LLM and Diffusion Model (DM), and traditional Convolutional Neural Network (CNN) tasks. The specific tasks are outlined as follows:

\begin{itemize}
    \item \textbf{Large Language Model(Transformer Decode Only)} We pre-train GPT-2~\cite{radford2019language} on OpenWebText~\cite{Gokaslan2019OpenWeb}. We profile the optimizer state tensors' memory usage and optimizer step time for GPT-2-XL(1.5B) and LLM2-7B.
    \item \textbf{Vision Transformer} We investigate the Vision Transformer (ViT/S-16)~\cite{dosovitskiy2021imageworth16x16words} on the ImageNet-1k dataset~\cite{deng2009imagenet} for image classification tasks.  We profile the optimizer state tensors' memory usage and optimizer step time for ViT-H/14.
    \item \textbf{Parameter-Efficient Fine-Tuning (PEFT) LoRA} Furthermore, we explore Parameter-Efficient Fine-Tuning (PEFT) tasks for GPT-2 LoRA\cite{hu2021lora} fine-tuning on the E2E~\cite{novikova2017e2e} dataset and Diffusion Model fine-tuning to capture visual concepts. For image classification, we report the top-1 validation accuracy, while for Large Language Model (LLM) fine-tuning tasks, to evaluate the results of the fine-tuning, we report metrics such as BLEU~\cite{papineni2002bleu}, NIST~\cite{doddington2002automatic}, MET~\cite{banerjee2005meteor}, ROUGE-L~\cite{lin2004rouge}, and CIDEr~\cite{vedantam2015cider}. For all these metrics, higher scores indicate better performance. Additionally, we perform qualitative evaluations for the Diffusion Model (DM) fine-tuning task.
    \item \textbf{Convolutional Neural Networks (CNNs).} We study ResNet-18 (11M parameters) on the CIFAR-10 dataset, and architectures from NATS-Bench~\cite{dong2021nats} on CIFAR-10, CIFAR-100, and ImageNet16-120~\cite{krizhevsky2009learning, chrabaszcz2017downsampled}. All these are image classification tasks, and we report the top-1 test accuracy as the evaluation metric. 
\end{itemize}

\subsection{LLM Pre-train}\label{sec:exp:pretrain_llm}
\textbf{Setups.} 
We pre-train GPT-2-Small (125M)~\cite{radford2019language} on OpenWebText~\cite{Gokaslan2019OpenWeb}. We compare SGD-SaI with AdamW~\cite{loshchilov2019decoupled} and Adam-mini~\cite{zhang2024adamminiusefewerlearning}. We follow the same settings as described in the previous study~\cite{zhang2024adamminiusefewerlearning}. We analyse the loss metrics for each optimizer. 

For large-scale LLMs, we provide profiling results focusing on memory usage and wall-clock time during the optimizer step for GPT-2-XL (1.5B parameters) and Llama-2 (7B parameters). Due to resource constraints, these results are limited to the optimizer step time and do not encompass full training runs.
We compare SGD-SaI with SGDM, AdamW, Adam~\cite{kingma2014adam}, Adam-mini, and Prodigy~\cite{mishchenko2023prodigy}.
The reported metrics include memory usage of the state tensors and the time costs associated with the optimizer steps. All results were obtained using a single NVIDIA A100 (80GB).

\textbf{Results.} 
Figure~\ref{fig:optimizer_metrics_comparison} compares optimizers (AdamW, Adam-mini, and SGD-SaI) during pre-training of GPT-2-Small across multiple metrics. While SGD-SaI demonstrates a slightly slower initial convergence speed compared to the Adam family optimizers due to its design, it achieves superior final convergence with a lower training loss (outperforming Adam-mini by 0.13). Similarly, validation loss shows a marginal improvement, with SGD-SaI reducing it by 0.03 compared to Adam-mini.

\textbf{Efficiency.}
\textit{For the GPT-2-Small pre-training task.} Regarding update speed, SGD-SaI demonstrates a significant advantage, being three times faster than Adam-mini in parameter updates and outperforming AdamW. Furthermore, the memory efficiency of SGD-SaI is noteworthy—it consumes only half the memory required by AdamW while maintaining performance comparable to Adam-mini, which employs intricate partitioning strategies. Unlike Adam-mini, which requires complex parameter partitioning (eg. users need to manually transform the Pytorch default partitions like the combined QKV block into separate Q, K and V blocks.), SGD-SaI achieves similar or better results using the default PyTorch partitioning, highlighting its simplicity and efficiency.
\textit{For the untrained models.} By design, the state tensors for Adam and AdamW are approximately twice the size of the gradient, while Prodigy’s state tensors are roughly four times larger. In contrast, SGD-SaI has state tensors of the same size as standard SGDM. This effectively reduces memory usage by up to 75\% compared to Prodigy and by 50\% compared to Adam(W). The detailed discussion can be found in the Appendix~\ref{sec:sup:optimizer_breakdown}. As shown in Table~\ref{table:llm_profiled_performance}, SGD-SaI maintains a manageable memory footprint, enabling it to work with large models like Llama-2 (7B) without running into out-of-memory (OOM) errors. In contrast, other optimizers, such as AdamW and Prodigy, exceed available memory limits at this model size, highlighting the scalability challenges posed by memory-intensive optimizers when dealing with long context lengths in LLMs.
Adam-Mini requires a partitioning strategy for different parameter groups while adjusting the learning rate adaptively at each time step. This increases memory usage and computational cost as different groups can not update simultaneously. For models larger than 1 billion parameters, the performance gains from Adam-Mini decrease by approximately 45\%, while the reduction achieved with SGD-SaI remains around 50\%.

\begin{table}[!ht]
\centering
\small 
\setlength{\tabcolsep}{6pt} 
\renewcommand{\arraystretch}{1.3} 
\resizebox{0.9\columnwidth}{!}{
\begin{tabular}{l|l|l|l}
\hline
Model & Method & State Mem (GB) & Wall Time (ms) \\ \hline
\multirow{6}{*}{GPT2-1.5B} 
    & SGDM        & 5.93  & 41.0 $\pm$ 12.0 \\
    & AdamW       & 11.86  & 138.0 $\pm$ 6.0 \\
    & Adam        & 11.86  & 145.0 $\pm$ 7.0 \\
    & Prodigy     & 23.72  & 360.0 $\pm$ 45.0 \\
    & Adam-Mini   & 6.52  & 223.0 $\pm$ 2.0 \\
    & \textbf{SGD-SaI(ours)}    & 5.93                        & 68.0 $\pm$ 21.0 \\ \hline
\multirow{6}{*}{Llama2-7B} 
    & SGDM        & 25.15                       & 100.0 $\pm$ 20.0 \\
    & AdamW       & 49.48                       & OOM \\
    & Adam        & 49.48                       & OOM \\
    & Prodigy     & 98.96                       & OOM \\
    & Adam-Mini   & 27.21                       & 421.0 $\pm$ 22.0 \\
    & \textbf{SGD-SaI(ours)}     & 25.15                       & 180.0 $\pm$ 30.0 \\ \hline
\end{tabular}
}
\caption{The efficiency metrics of various models with different optimizers were evaluated using an A100-80GB GPU. The table above summarizes the results, which include the tensor memory usage and wall-clock time (optimization step time measured in milliseconds) for each model-optimizer configuration. For the large language models (LLMs), experiments were conducted with a context length of 1024 and a batch size of 1. All models were profiled in full (FP32) precision.}
\label{table:llm_profiled_performance}
\end{table}

\subsection{ViT Pre-train}\label{sec:exp:pretrain_vit}
\textbf{Setups.} We pre-train ViT-S/16~\cite{dosovitskiy2021imageworth16x16words} on the ImageNet1k dataset~\cite{deng2009imagenet} for the image classification task.
We compare SGD-SaI with AdamW~\cite{loshchilov2019decoupled} as well as popular optimizers including SGDM, Adam~\cite{kingma2014adam}, Adam-mini~\cite{zhang2024adamminiusefewerlearning} and Prodigy~\cite{mishchenko2023prodigy}.
After conducting a grid search within the same hyperparameter range, we compare the optimiser results. We report the peak and mean of the top-1 validation accuracy to evaluate their generalisation ability and sensitivity to hyperparameter changes. Detailed hyperparameters are in Appendix~\ref{sec:sup:vit_gs_details}.

Due to the intensive computational power requirements for the ViT variants during the grid search, we cannot provide the complete training results. Instead, we follow the same procedure outlined in Section~\ref{sec:exp:pretrain_llm} and present only the profiling results regarding memory usage and wall-clock time during the optimizer step for ViT-S/16 and ViT-H/14. Additionally, we compare SGD-SaI with SGDM, AdamW, Adam, Adam-mini, and Prodigy.
All results were obtained using a single NVIDIA A100 with 80GB of memory.

\textbf{Results.} 
We report peak performance under the best hyperparameters, averaging results over three random seeds, and present the mean and standard deviation (see Table~\ref{table:vit_results}). Our simple re-scaling strategy significantly boosts SGDM's performance from 63.80 to 72.92, nearly matching AdamW's 73.04. Meanwhile, the recent SOTA optimizer Prodigy achieves a slightly higher peak at 73.24, though it requires additional one-time memory usage. We will discuss these results further in later sections. Notably, our approach achieves the lowest standard deviation (0.07) across three random seeds, compared to Prodigy’s second-lowest at 0.21, highlighting the stability of our method during training.

In addition, we examine average performance across the hyperparameter search grid using a rest setting that deviates from the best hyperparameters as a tweaked version. Under these conditions, we observe that most previous methods, including AdamW, struggle significantly, leading to dramatic drops in average performance. For example, AdamW, despite being an update over SGD intended to improve robustness to hyperparameters, achieves only 37.21 with a standard deviation of 35.43. In contrast, our method maintains overall performance, achieving an average of 57.55 with a much lower variance (standard deviation of 18.46). Prodigy, a parameter-free optimizer not designed to adjust learning rate and weight decay, fails to converge when these hyperparameters are modified; thus, we exclude it from this part of the comparison for fairness.

\begin{table}[!ht]
\resizebox{0.96\columnwidth}{!}{
\begin{tabular}{lcc}
\hline
\textbf{Optimizer} & \textbf{Peak@top1 (\%)}      & \textbf{Avg@top1 (\%)}        \\ \hline
SGDM               & 63.80 $\pm$ 0.35             & 14.33 $\pm$ 19.38             \\
Adam               & 61.56 $\pm$ 0.93             & 20.93 $\pm$ 22.05             \\
Adam-Mini           & 72.29 $\pm$ 0.43             & 36.65 $\pm$ 35.39             \\
AdamW              & \underline{73.04} $\pm$ 0.31 & \underline{37.21} $\pm$ 35.43 \\ \hline
Prodigy            & \textbf{73.24} $\pm$ 0.21    & N/A                           \\ \hline
SGD-SaI(Ours)    & 72.92 $\pm$ \textbf{0.07}             & \textbf{57.55} $\pm$ \textbf{18.46}    \\ \hline
\end{tabular}
}
\caption{Comparison of peak and average top-1 validation accuracy on ImageNet-1k for ViT-S/16 trained from scratch. Each optimizer's performance is evaluated over a hyperparameter search space, reporting the highest accuracy (Peak@top1) and the average accuracy (Avg@top1) across all trials. Results are averaged over three seeds, with standard deviations for statistical analysis. Our method achieves significantly higher robustness to hyperparameter variations, maintaining a high average performance (57.55\%) and outperforming other optimizers by at least 20\%.}
\label{table:vit_results}
\end{table}

Our method demonstrates superior robustness and effectiveness in training ViT-S/16 models from scratch on ImageNet-1k, outperforming previous optimizers across peak and average performance metrics. While alternative optimizers like AdamW and Prodigy achieve high peak accuracy, their performance drops significantly under hyperparameter variations, highlighting their sensitivity. In contrast, our approach maintains a stable and high average accuracy across diverse hyperparameter settings with minimal standard deviation. It underscores its resilience to hyperparameter tuning and potential for more efficient and reliable model training in real-world applications. This stability makes our method particularly suitable for scenarios where hyperparameter tuning is constrained, offering a consistent and robust solution for training large models.

Although our method does not utilize adaptive gradient adjustments, it achieves a stable and steady learning pace, ultimately reaching a comparable performance, as shown in Figure~\ref{fig:vit_loss_acc_across_time}. Empirically, this stability arises from our preconditioned learning rate, which ensures each step is well-controlled and converges reliably with sufficient training steps. In contrast, Adam-family optimization methods often achieve faster convergence but are prone to being trapped in suboptimal minima due to their aggressive adaptivity.

\textbf{Efficiency.} 
As shown in Table~\ref{table:vit_profiled_performance}, our method achieves a wall-clock time for optimizer steps comparable to SGDM, while being significantly faster than Adam-mini, Adam(W), and Prodigy. We must note that we present the wall clock time for each optimizer step rather than the total runtime. We did not rely on the grid search results to report the total runtime because all grid search experiments were conducted on a cluster. Due to complex factors, such as the cluster's I/O bottleneck and network congestion, distributed training can be considerably slowed down. We chose to maintain the same settings and device while profiling the LLM. Regarding memory usage, similar trends were observed during the pre-training of GPT-2 Small. For example, the ViT-H/14-0.66B model uses only 2.42 GB of memory with SGD-SaI, compared to 4.86 GB with AdamW and 9.70 GB with Prodigy. Our method reduces memory consumption by 50\% compared to Adam(W) and by 75\% compared to Prodigy. Further empirical analysis can be found in App.~\ref{sec:sup:optimizer_breakdown}.

\begin{table}[!ht]
\centering
\small 
\setlength{\tabcolsep}{6pt} 
\renewcommand{\arraystretch}{1.3} 
\resizebox{0.9\columnwidth}{!}{
\begin{tabular}{l|l|l|l}
\hline
Model & Method & State Mem (GB) & Wall Time (ms) \\ \hline
\multirow{6}{*}{ViT-S/16(0.0229B)} 
    & SGDM        & 0.08  & 7.9 $\pm$ 0.3 \\
    & AdamW       & 0.17  & 45.0 $\pm$ 8.0 \\
    & Adam        & 0.17 & 50.0 $\pm$ 1.5 \\
    & Prodigy     & 0.33  & 78.0 $\pm$ 0.0 \\
    & Adam-Mini   & 0.08  & 84.0 $\pm$ 5.0 \\
    & \textbf{SGD-SaI(ours)}   & 0.08 & 12.4 $\pm$ 0.2 \\ \hline
\multirow{6}{*}{ViT-H/14(0.66B)} 
    & SGDM        & 2.42  & 40.0 $\pm$ 1.0 \\
    & AdamW       & 4.86  & 124.0 $\pm$ 4.0 \\
    & Adam        & 4.86  & 127.0 $\pm$ 2.0 \\
    & Prodigy     & 9.70  & 260.0 $\pm$ 3.0 \\
    & Adam-Mini   & 2.54  & 220.0 $\pm$ 20.0 \\
    & \textbf{SGD-SaI(ours)}    & 2.42                        & 54.0 $\pm$ 13.0 \\ \hline
\end{tabular}
}
\caption
{We maintain the same settings as in Table~\ref{table:llm_profiled_performance}. We ensure a comparable memory footprint to SGDM, while keeping the optimizer step time controlled, resulting in a performance that is \textbf{4-6 x faster than} Adam-mini.}
\label{table:vit_profiled_performance}
\end{table}

\begin{figure*}[!ht]
    \centering
    \includegraphics[width=0.9\textwidth]{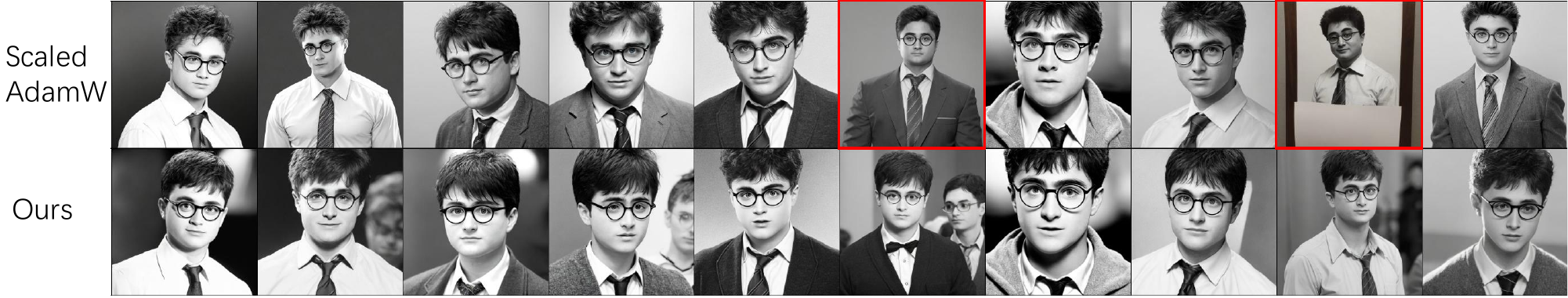}     
    \caption{Generation results for prompt “a pencil sketch of ⟨Vpotter⟩” by Mix-of-Show model with \textbf{scaledAdamW optimizers(up)} and \textbf{our optimizer(down)}. Our method generates photos that better capture the prompt and align with visual concepts from training samples; At the same time, previous SOTA-scaled AdamW has some significant bad cases that do not follow the prompt, we marked them with a red bounding box.}
    \label{fig:mix_of_show}
    \vspace{-3mm}
\end{figure*}

\begin{figure*}[!ht]
    \centering
     \includegraphics[width=.32\textwidth]{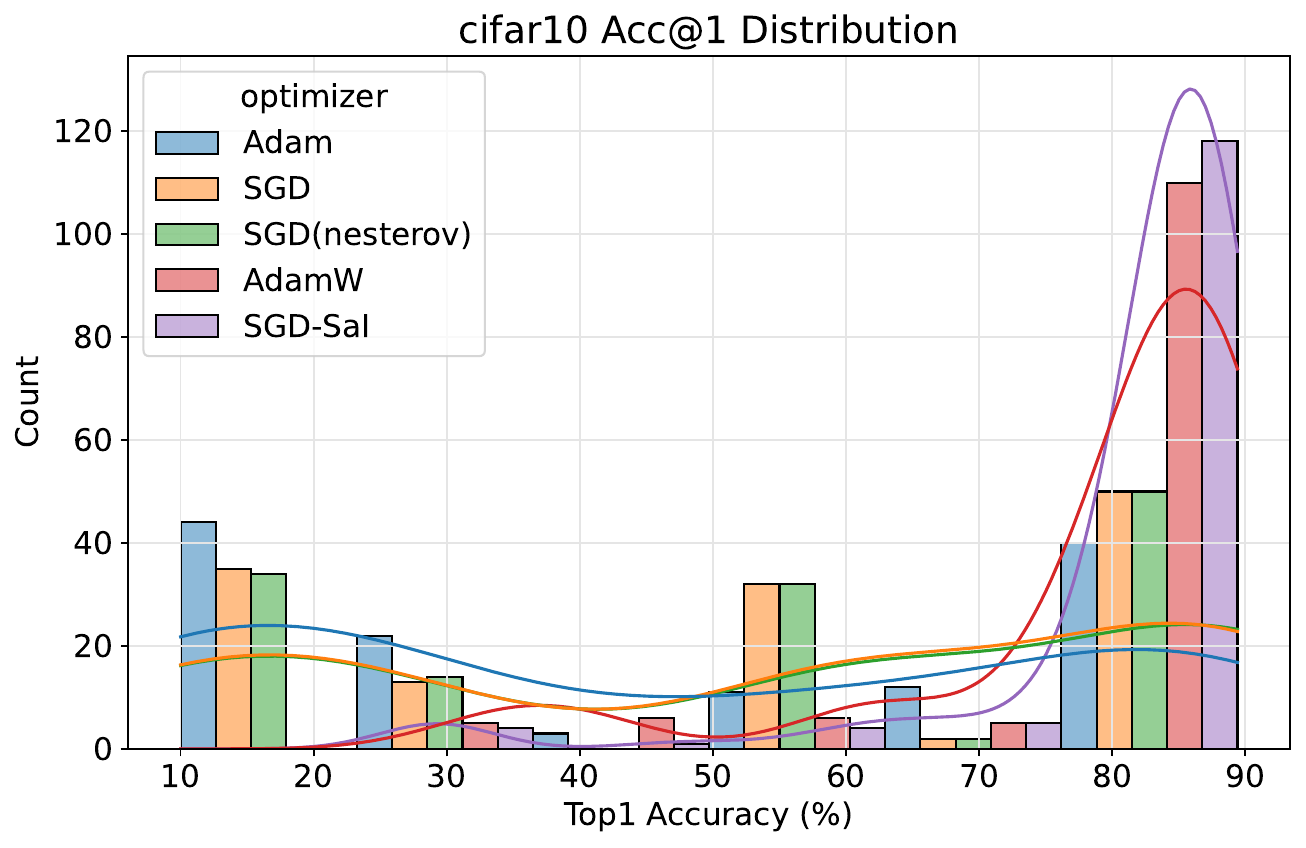}
     \includegraphics[width=.32\textwidth]{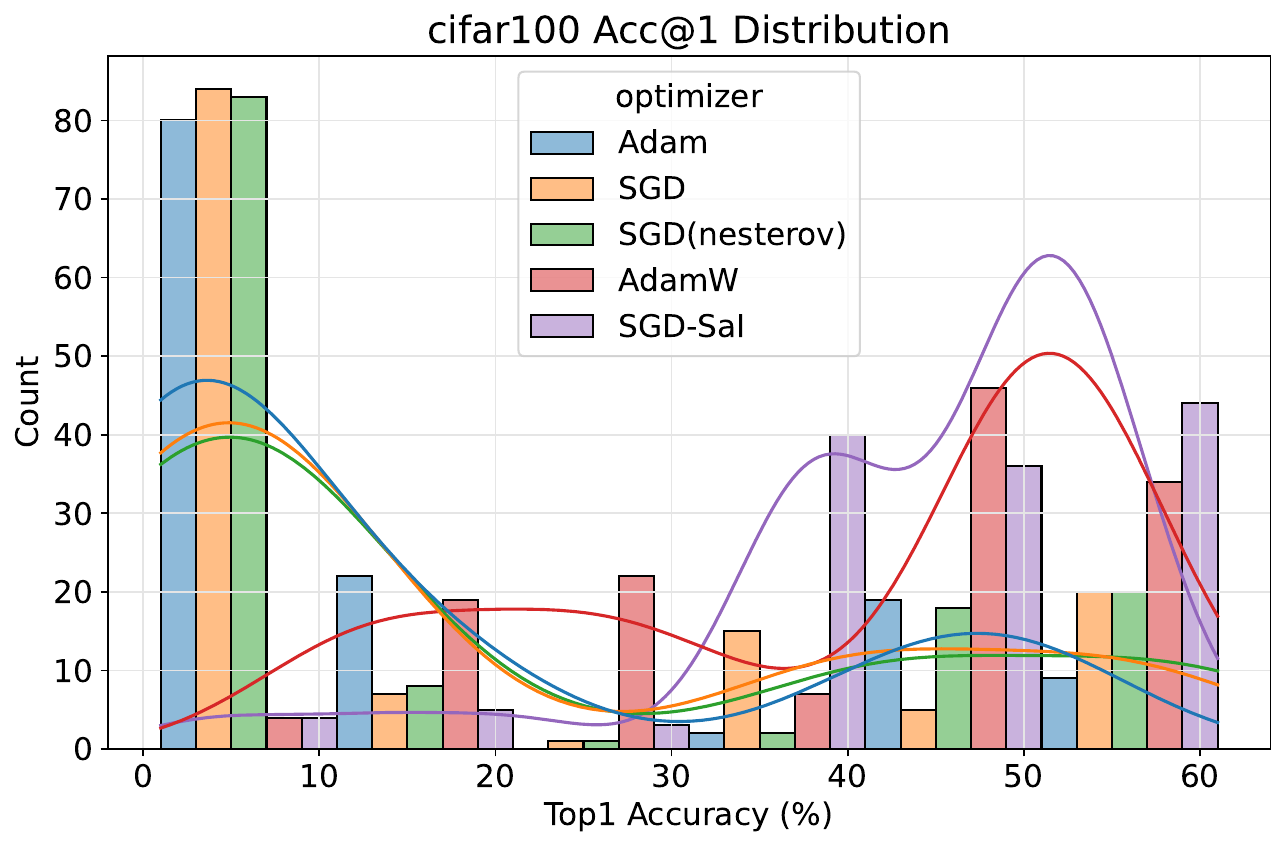}
     \includegraphics[width=.32\textwidth]{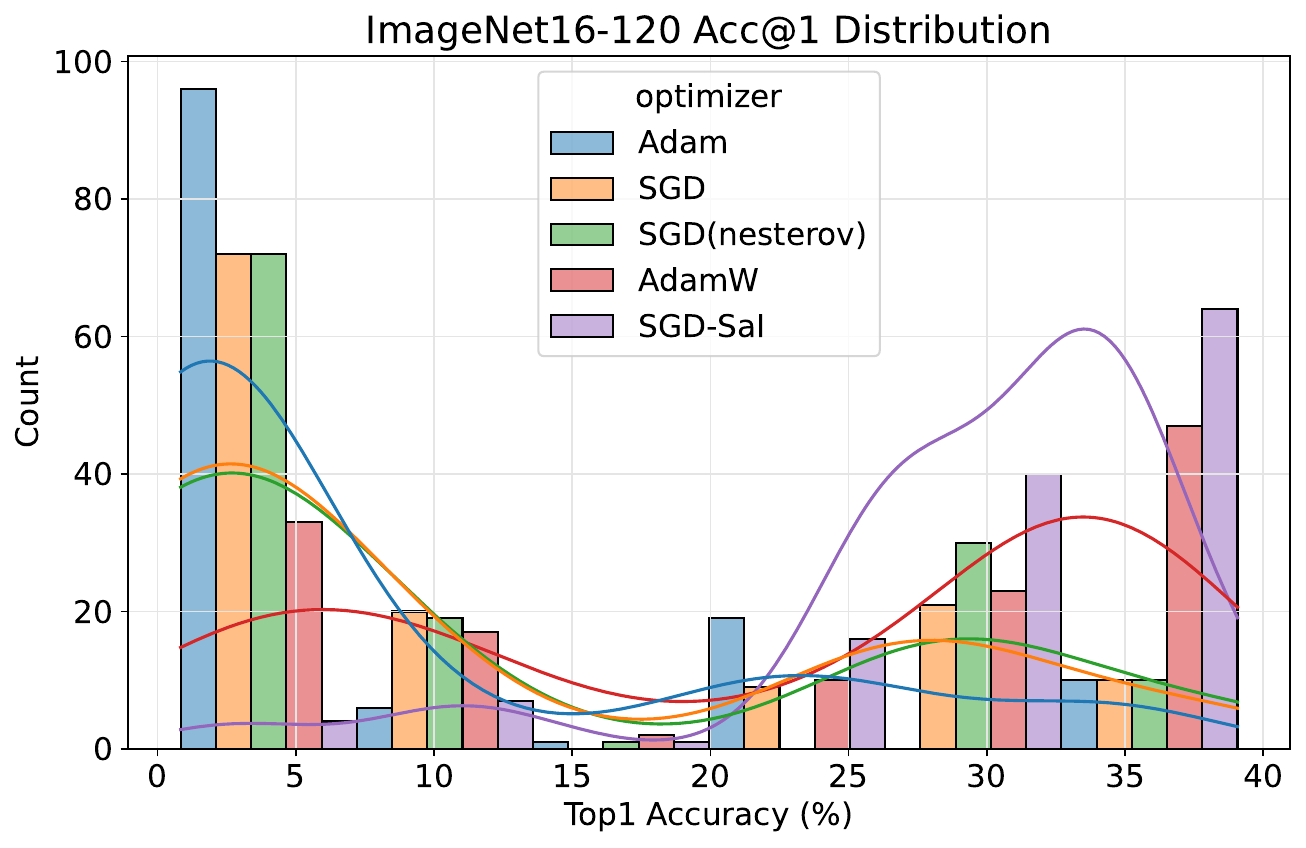}
    \caption{These figures show the accuracy distributions of eleven architectures trained on different optimizers using the same hyperparameter candidates.  This row presents the top-1 evaluation accuracy distributions on CIFAR-10, CIFAR-100 and ImageNet16-120. The curves in those histograms are the results of kernel density estimation (KDE).}
    \label{fig:fig_sss_performance_distribution_histogram}
\end{figure*}

\subsection{Parameter Efficient Fine Tuning: PEFT}\label{sec:exp:finetune_peft}
We primarily consider PEFT tasks on LLM fine-tuning and the Diffusion Model fine-tuning. 

\subsubsection{LLMs Parameter Efficient Fine-tuning}

\textbf{Setup.} 
We fine-tune the GPT-2 model using the E2E dataset ~\cite{novikova2017e2e}. The current state-of-the-art (SOTA) methods include scaled-SGD and scaled-AdamW from ~\cite{zhang2024riemannian}, which adjust the learning rates for parameters $A$ and $B$ with a Riemannian preconditioner. Our primary comparison is between SGD-SaI and these methods, along with Adam-mini. To evaluate the results of the fine-tuning, we report metrics such as BLEU~\cite{papineni2002bleu}, NIST~\cite{doddington2002automatic}, MET~\cite{banerjee2005meteor}, ROUGE-L~\cite{lin2004rouge}, and CIDEr~\cite{vedantam2015cider}. For all these metrics, higher scores indicate better performance.

\textbf{Results.} 
We adopted the same experiment setting to investigate whether our methods suit LoRA Training~\cite{hu2021lora}. We set the default learning rate to \(1e-3\) and weight decay to \(1e-2\). Empirically, we observe that SGD-SaI outperforms previous state-of-the-art (SOTA) scaled optimizers and unscaled ones. Table~\ref{tab:lora_scores} presents surprising results regarding the final scores for LoRA fine-tuning of the GPT-2 medium model with a rank of 4 on the E2E natural language generation tasks. With this simple precondition on SGDM, our method performs significantly better than the previous SOTA strategy using rescaled SGD. Furthermore, our approach exhibits a substantial improvement over AdamW in fine-tuning the GPT-2 architecture, even without meticulous tuning and searching for hyperparameters.

\begin{table}[!ht]
\centering
\resizebox{\columnwidth}{!}{
\begin{tabular}{l|c|c|c|c|c}
\hline
\textbf{Method} & \textbf{BLEU} & \textbf{NIST} & \textbf{MET} & \textbf{ROUGE-L} & \textbf{CIDEr} \\ \hline
SGD\textsubscript{\(r=4\)} & 66.6 & 8.54 & 44.2 & 68.2 & 2.32 \\ 
scaled SGD\textsubscript{\(r=4\)} & 69.2 & 8.71 & 46.3 & 70.9 & 2.48 \\ 
AdamW\textsubscript{\(r=4\)} & 68.9 & 8.69 & 46.5 & 71.3 & 2.51 \\
Adam-Mini\textsubscript{\(r=4\)} & 68.7 & 8.66 & 46.3 & 71.1 & 2.50 \\
scaled AdamW \textsubscript{\(r=4\)} & 69.6 & 8.77 & 46.6 & 71.8 & 2.52   \\ \hline
SGD-SaI (ours)\textsubscript{\(r=4\)} & \textbf{69.9} & \textbf{8.81} & \textbf{46.7} & \textbf{72.1} & \textbf{2.53} \\
\end{tabular}
}
\caption{This table presents scores for LoRA fine-tuning of GPT-2 medium model on E2E Natural Language Generation (NLG) challenge with different optimizers. SGD-SaI outperforms all scaled and unscaled optimizers on all evaluation metrics. In particular, our method closes the performance gap between SGD and AdamW and reveals its effectiveness in performing block-wise scaling.}

\label{tab:lora_scores}
\end{table}

\subsubsection{DMs Parameter Efficient Fine-tuning}

\textbf{Setup.} 
Using the diffusion model, we extend our experiments to include LoRA fine-tuning on image generation tasks. Specifically, we utilize the ChilloutMix model to address real-world concepts, following the same approach outlined in Mix-of-show \cite{gu2024mix, zhang2024riemannian}. Additionally, we compare our method with the state-of-the-art (SOTA) optimized approach using scaled-AdamW~\cite{zhang2024riemannian}. To evaluate the images generated by the diffusion model, we conduct a qualitative assessment to determine which method captures visual concepts more effectively.

\textbf{Results.}
Face generation is a challenging task; the model should understand the visual concept of a specific person's face based on its prompt text. Here, we set the learning rate as default 0.1, a large enough default value. We observed as Fig.~\ref{fig:mix_of_show}, even without carefully tuning the learning rate, our scaled methods have shown a significantly better ability to capture the visual concept of \textbf{potter} than the previous SOTA scaled approach scaled-AdamW~\cite{zhang2024riemannian}. It should be verified that our optimizer has better parameters robustness on training and leads to better convergence in final performance; this should be an essential benefit for the practical use of the optimizer.

\subsection{Convolutional Neural Network(CNN)} \label{sec:exp:pretrain_cnn} 

\textbf{Setup.} 
We follow a similar approach to Section~\ref{sec:exp:pretrain_vit} for evaluating CNN models. A grid search is performed on ResNet18~\cite{he2015deepresiduallearningimage} using the CIFAR-10 dataset, and across various architectures from NATS-Bench~\cite{dong2021nats} on CIFAR-10, CIFAR-100, and ImageNet16-120~\cite{krizhevsky2009learning, chrabaszcz2017downsampled}. All tasks involve image classification. We compare SGD-SaI with traditional optimizers (SGD and Adam-family) and report top-1 test accuracy. Details of the grid search experiments are provided in Appendix~\ref{see:sup:cnn_gs_details}.

\textbf{Results.} 
Figure~\ref{fig:fig_vit_cnn_performance_3d_surface} (left graph) presents the performance of ResNet18. Our method achieves a peak accuracy of 95.36\%, which not only surpasses that of Adam(W) and SGD but also shows greater stability. 
In addition, we evaluated a range of search spaces, including datasets such as CIFAR-10, CIFAR-100, and ImageNet16-120, as well as architectures of varying sizes. We conducted a grid search across eleven architectures, testing three learning rates and four weight decay values. The distribution of top-1 accuracies is illustrated in Fig.~\ref{fig:fig_sss_performance_distribution_histogram}, which demonstrates the stability of our method across different architectures and hyperparameter settings. 
Our approach results in models with lower standard deviations and higher mean accuracies, indicating enhanced stability and generalization. These findings highlight the robustness of our method across various CNN architectures.

\section{Conclusion}
In summary, our results demonstrate that simply applying selective learning rate scaling at initialization (SGD-SaI) can unlock performance comparable to—if not better than—leading adaptive gradient methods like AdamW, all while retaining the simplicity and efficiency of SGDM. By leveraging g-SNR to guide parameter group scaling, SGD-SaI not only mitigates early training imbalances but also substantially reduces optimizer memory overhead, enabling more resource-efficient model training. Its robustness across a wide range of Transformer-based tasks, including ImageNet classification with ViT, GPT-2 pretraining, LoRA fine-tuning, and diffusion modelling, underscores its versatility and practicality. 


\section{Limitation}
While SGD-SaI demonstrates promising results across various Transformer-based tasks, our study is constrained by limited computational resources, preventing us from conducting large-scale pre-training on more extensive models such as Llama-2-7B. This remains an avenue for future research. However, to address the efficiency challenges of training larger models, we have performed detailed profiling of GPU memory usage and optimizer step speed on these architectures. These preliminary analyses indicate the potential scalability of SGD-SaI, but comprehensive evaluations on larger-scale models are necessary to establish its effectiveness and efficiency in such settings fully. Moreover, our methods ensure a steady and stable update during training, allowing the model to converge better in a given task with sufficient training steps. Thus, we might observe that the convergence speed is relatively lower than Adam's in the early stage of training; as our primary focus is to investigate the effectiveness of the SaI approach, we left the acceleration of convergence speed in future work.



\bibliography{main}
\bibliographystyle{icml2025}

\clearpage
\newpage
\newpage
\appendix


\section{More Experiments Details}

\subsection{Details for ViT Experiments}\label{sec:sup:vit_gs_details}
In this section, we will list list the settings of the experiment regarding to Section~\ref{sec:exp:pretrain_vit}.

\textbf{Hyperparameter Settings:} We start by following the settings in~\cite{beyer2022betterplainvitbaselines, steiner2022trainvitdataaugmentation, big_vision};  Specifically, we include Nesterov-SGD as a baseline, offering better performance than naive SGD. All optimizers are tested using a grid search within the same hyperparameter ranges: learning rate \( lr \in \{0.1, 0.01, 0.001, 0.0001\} \) and weight decay \( wd \in \{0.01, 0.001, 0.0001\} \).

\subsection{Details for CNN Experiments}\label{see:sup:cnn_gs_details}
In this section, we will list the settings of the experiment regarding to Section~\ref{sec:exp:pretrain_cnn}.

\textbf{Models and Datasets:}  We follow the same settings and train some CNN-based architectures proposed in NATS-Benchmark ~\cite{dong2021nats}. We test the optimizers on CIFAR-10/CIFAR-100 ~\cite{krizhevsky2009learning} and ImageNet16-120 ~\cite{chrabaszcz2017downsampled}. Based on the NATS-Benchmark work, we test different sizes of architectures. Here, we select ten architectures with top-10 validation accuracy and one architecture with bottom-1 validation accuracy in terms of different datasets and training epochs to present. 

\textbf{Hyperparameter Settings:} The optimal learning rate and weight decay are chosen by performing the grid search. The learning rate and weight decay are selected from $\eta \in \{0.1, 0.01, 0.001\}$ and $\lambda \in \{0.5, 0.05, 0.005, 0.0005\}$, respectively. We use the same cosine annealing scheduler on three datasets without learning rate warmup. We use the same data augmentation methods and set the batch size to 256 for all datasets. The experiments are designed to run for full training without early stopping. There is no linear scaling on the initial weight decay either since we are doing the grid search within a feasible range. The seed is only {777} which is the same seed reported by NATS-Benchmark on Size Search Space. The original NATS-Benchmark were produced using SGD with a fixed learning rate 0.1 and weight decay 0.0005 and the default setting of the Nesterov Momentum. For fair comparison, we apply the same grid search policy for SGD, as the baseline with or without Nesterov Momentum. 

\textbf{Results:} The performance of various architectures has been represented as histograms showing top-1 accuracy on different datasets. Fig.~\ref{fig:fig_sss_performance_distribution_histogram} demonstrates that our method outperforms other optimizers in terms of evaluation accuracies within the same hyperparameter search space.

\subsection{Extra Results for ResNet18 on CIFAR10}

As a classic model of the CNNs, we also conduct the grid search on ResNet18 as an extended experiment. 

\textbf{Models and Datasets:} We follow the similar setting in the Section~\ref{sec:exp:pretrain_cnn}. We particularly choose the CIFAR-10~\cite{krizhevsky2009learning} as the dataset we test on. We test on the classic ResNet18 model.

\textbf{Hyperparameter Settings:} Since we are focusing on a single model with one dataset—unlike the NAST-Benchmark CNN experiments discussed in Section~\ref{sec:exp:pretrain_cnn}—we are scaling up our search by exploring a wider range of learning rates and weight decays. The learning rates are chosen from the set \(\eta \in \{0.1, 0.01, 0.001, 0.0001\}\) and the weight decays from \(\lambda \in \{0.5, 0.05, 0.005, 0.0005, 0.00005\}\). We will repeat our grid search three times using three different random seeds. The random seeds used for the experiments are \{42, 888, 999\}.
We opted for a step learning rate scheduler rather than a cosine annealing scheduler to test our method's resilience to different learning rate scheduling policies. The learning rate will decrease by a factor of 10 every 80 epochs, with a total of 200 epochs for training.
Our data augmentation methods remain consistent, with a batch size set to 128. These experiments will run for the entire training duration without early stopping, and there will be no linear scaling applied to the initial weight decay, as we are conducting a grid search within a reasonable range. 
The distribution of accuracies averaged over the three seeds for each hyperparameter combination is depicted in Fig.\ref{fig:fig_vit_cnn_performance_3d_surface}. The best performance of each optimizer, along with the optimal learning rate and weight decay, is annotated with red numbers on the graph. For simplicity, we are only testing the Stochastic Gradient Descent with Momentum (SGDM) as the baseline. The momentum is set to the default value of 0.9, consistent with both the Adam(W) optimizer and our method to ensure fair comparison.

\textbf{Results: } The grid search results are shown in the Fig.~\ref{fig:fig_vit_cnn_performance_3d_surface}. Not only does our method converge better (ours 95.36\% v.s. SGDM 95.26\%), but it also demonstrates greater resilience to changes in hyperparameters. This means the performance is less likely to downgrade compared to SGDM.

All the experiments in this paper were conducted using various types of GPUs, including NVIDIA GeForce RTX 3090, NVIDIA A100 PCIe 40GB, and NVIDIA A100 80GB. To ensure consistent experimental conditions, each experiment was conducted using only one GPU type.

\begin{figure}[t]
\begin{algorithm}[H]
    \caption{SGD}\label{algorithm:sgd}
    \footnotesize
    \begin{algorithmic}[1]
        \Require $t$ (step), $\eta$ (lr), $\theta^i$ (i-th params), $L(\theta)$ (loss function), $\lambda$ (weight decay), $\mu$ (momentum), $\tau$ (dampening),  $maximize$ 
        
        \Repeat
        \For{$t\gets 1$}
            \State $g^i_t \gets \nabla L(\theta^i_{t-1})$
            \State 
            \State \text{/* do weight decay */}
            \State \text{$g^i_t \gets g^i_t + \lambda \theta^i_{t-1}$}
            \State
            \State \text{/* do momentum */}
            \If{$\mu \neq 0$ \textbf{and} $t > 1$}
                \State {$m^i_t \gets \mu m^i_{t-1} + (1 - \tau) g^i_t$}
            \Else
                \State {$m^i_t \gets g^i_t$}
            \EndIf
            \State
            \If{$maximize$}
                \State $\theta^i_t \gets \theta^i_{t-1} + \eta m^i_t$
            \Else
                \State $\theta^i_t \gets \theta^i_{t-1} - \eta m^i_t$
            \EndIf
        \EndFor
        \Until {epochs end}
    \end{algorithmic}
\end{algorithm}
\end{figure}

\begin{figure}[t]
\begin{algorithm}[H]
    \caption{Adam}\label{algorithm:adam}
    \begin{algorithmic}[1]
        \Require $t$ (step), $\eta$ (lr), $\theta^i$ (i-th params), $L(\theta)$ (loss function), $\lambda$ (weight decay), $\beta_1, \beta_2$ (betas),  $maximize$ 
        
        \Repeat
        \For{$t\gets 1$}
            \If{$maximize$}
                \State $g^i_t \gets -\nabla L(\theta^i_{t-1})$
            \Else
                \State $g^i_t \gets \nabla L(\theta^i_{t-1})$
            \EndIf
            \State 
            \State \text{/* do weight decay */}
            \State \text{$g^i_t \gets g^i_t + \lambda \theta^i_{t-1}$}
            \State
            \State \text{/* do momentum */}
            \If{$t > 1$}
                \State {$m^i_t \gets \beta_1 m^i_{t-1} + \text{$(1-\beta_1)$} g^i_t $}
                \State $v^i_t \gets \beta_2 v^i_{t-1} + (1-\beta_2) (g^i_t)^2 $
            \Else
                \State \text{$m^i_t \gets (1-\beta_1) g^i_t $}; $v^i_t \gets (1-\beta_2) (g^i_t)^2 $
            \EndIf
            \State{$\hat{m^i_t} \gets \frac{m^i_t}{1-\beta_1^t}$};{$\hat{v^i_t} \gets \frac{v^i_t}{1-\beta_2^t}$}
            \State
            
            \State $\theta^i_t \gets \theta^i_{t-1} - \text{$ \frac{\eta}{\sqrt{\hat{v^i_t}} + \epsilon}m^i_t $} $

        \EndFor
        \Until {epochs end}
    \end{algorithmic}
\end{algorithm}
\end{figure}

\begin{figure}[t]
\begin{algorithm}[H]
    \caption{AdamW}\label{algorithm:adamw}
    \begin{algorithmic}[1]
        \Require $t$ (step), $\eta$ (lr), $\theta^i$ (i-th params), $L(\theta)$ (loss function), $\lambda$ (weight decay), $\beta_1, \beta_2$ (betas),  $maximize$ 
        
        \Repeat
        \For{$t\gets 1$}
            \If{$maximize$}
                \State $g^i_t \gets -\nabla L(\theta^i_{t-1})$
            \Else
                \State $g^i_t \gets \nabla L(\theta^i_{t-1})$
            \EndIf
            \State \text{/* do momentum */}
            \If{$t > 1$}
                \State {$m^i_t \gets \beta_1 m^i_{t-1} + \text{$(1-\beta_1)$} g^i_t $}
                \State $v^i_t \gets \beta_2 v^i_{t-1} + (1-\beta_2) (g^i_t)^2 $
            \Else
                \State \text{$m^i_t \gets (1-\beta_1) g^i_t $}; $v^i_t \gets (1-\beta_2) (g^i_t)^2 $
            \EndIf
            \State{$\hat{m^i_t} \gets \frac{m^i_t}{1-\beta_1^t}$};{$\hat{v^i_t} \gets \frac{v^i_t}{1-\beta_2^t}$}
            \State
            
            \State{\text{/* do weight decay */}}
            \State \text{$\theta^i_t \gets \theta^i_{t-1} - \lambda \eta \theta^i_{t-1}$}
            \State
            \State $\theta^i_t \gets \theta^i_{t-1} - \text{$ \frac{\eta}{\sqrt{\hat{v^i_t}} + \epsilon}m^i_t $} $

        \EndFor
        \Until {epochs end}
    \end{algorithmic}
\end{algorithm}
\end{figure}

\section{Optimizer Analysis} \label{sec:sup:optimizer_breakdown}
This section provides a supplementary analysis for Section~\ref{sec:results}. We will detail the optimizers and empirically estimate the lower boundary of the state tensors in memory.

\subsection{Break Down SGD}


Stochastic Gradient Descent (SGD) is faster than adaptive gradient methods primarily due to its simplicity. The key difference between SGD and these adaptive methods is that SGD uses a fixed learning rate, while adaptive methods adjust the learning rate dynamically for each parameter at each step. This adjustment can be done at the element level, as seen in Adam(W), or at the block level, like in Adam-mini.

The advantages of SGD can be summarized as follows: 
a. Runtime efficiency: It offers a fast and efficient iteration time for each optimization step.  
b. Memory efficiency: When using momentum, SGD requires only one instance of the gradients~\ref{algorithm:sgd} and incurs no additional memory overhead when momentum is not applied.

\subsection{Break Down Adam(W)}
The main difference between Adam and AdamW lies in how they apply weight decay. AdamW applies a direct penalty to the weights themselves, which is known as Decoupled Weight Decay, whereas Adam applies the penalty to the gradients at the outset, utilizing L2 Regularization. 
Both algorithms enhance adaptability by scaling the learning rate for each parameter individually. In every optimization step, the scaling ratios are recalculated by dividing the first-order moment by the square root of the second-order moment, both of which are maintained and updated in state tensors.
As illustrated in Alog.~\ref{algorithm:adam} (line 17 for Adam) and Alog.~\ref{algorithm:adamw} (line 14 for AdamW), the first-order moment $m$ and the second-order moment $v$ are stored in GPU memory throughout the entire training process.

Generally, the estimated minimum memory requirement for the state tensors in both Adam and AdamW is approximately twice the size of the gradient tensors. This is because $m$ and $v$ share the same shape as the corresponding gradient tensors.

\subsection{Break Down Adam-mini}

Adam-mini is a variant of the Adam optimizer. As shown in the Alog.~\ref{algorithm:adam_mini}, Adam-mini redesigns the adaptive update rules by using the mean of the squared gradients instead of the original squared gradients in most layers except for the embedding layer. This version of Adam reduces the number of learning rates to the number of blocks in each layer while keeping the update rules unchanged in the embedding layers. 
Therefore, the reduction in memory usage is influenced by the proportion of non-embedding parameters in the model. This limitation not only restricts the potential for memory savings but also incurs additional computational costs due to the extra operations (see Alog.~\ref{algorithm:adam_mini} Line 15, 16) needed when calculating the new $v$ compared to the original Adam algorithm.

Regarding the lower boundary of the state tensor memory, as mentioned in \cite{zhang2024adamminiusefewerlearning}, Adam-mini can reduce the memory used for the Adam optimizer's $v$ by at least 90\%. This results in a memory cost savings of approximately 45\% to 50\% compared to the original Adam. 

\begin{figure}[t]
\begin{algorithm}[H]
    \caption{Adam-mini}\label{algorithm:adam_mini}
    \begin{algorithmic}[1]
        \Require $t$ (step), $\eta$ (lr), $\theta^i$ (i-th params), $L(\theta)$ (loss function), $\lambda$ (weight decay), $\beta_1, \beta_2$ (betas),  $maximize$ 
        
        \Repeat
        \For{$t\gets 1$}
            \If{$maximize$}
                \State $g^i_t \gets -\nabla L(\theta^i_{t-1})$
            \Else
                \State $g^i_t \gets \nabla L(\theta^i_{t-1})$
            \EndIf
            \State
            \State \text{/* do momentum */}
            \If{$t > 1$}
                \State {$m^i_t \gets \beta_1 m^i_{t-1} + \text{$(1-\beta_1)$} g^i_t $}
                \If{$\theta^i_{t-1} \in embedding\_layer$}
                    \State $v^i_t \gets \beta_2 v^i_{t-1} + (1-\beta_2) (g^i_t)^2 $
                \Else
                    \State Divide $\theta^i_{t-1}$ into Q,K heads if needed.
                    \State $v^i_t \gets \beta_2 v^i_{t-1} + (1-\beta_2) Mean( (g^i_t)^2 ) $
                \EndIf
            \Else
                \State \text{$m^i_t \gets (1-\beta_1) g^i_t $}; $v^i_t \gets (1-\beta_2) (g^i_t)^2 $
            \EndIf
            \State{$\hat{m^i_t} \gets \frac{m^i_t}{1-\beta_1^t}$};{$\hat{v^i_t} \gets \frac{v^i_t}{1-\beta_2^t}$}
            \State
            
            \State{\text{/* do weight decay */}}
            \State \text{$\theta^i_t \gets \theta^i_{t-1} - \lambda \eta \theta^i_{t-1}$}
            \State
            \State $\theta^i_t \gets \theta^i_{t-1} - \text{$ \frac{\eta}{\sqrt{\hat{v^i_t}} + \epsilon}m^i_t $} $

        \EndFor
        \Until {epochs end}
    \end{algorithmic}
\end{algorithm}
\end{figure}

\subsection{Break Down Prodigy}

Prodigy is one of the most popular variants of the Adam optimizer, offering a new approach to calculating the step size. It alleviates the need for extensive learning rate tuning. While most of Adam's update rules remain unchanged, Prodigy introduces a new scaling ratio, denoted as $d$ (for D-Adaption), which adaptively adjusts the learning rate. To update the scaling ratio $d$ for each optimization step, Prodigy requires the maintenance of two additional tensors: the initial weight value $x_0$ and the denominator $s$, both of which share the same shape as the gradient.

As a result, the lower boundary for estimating the memory required by Prodigy's state tensor is approximately four times the size of the gradient by default. The majority of the memory for the tensor state is occupied by four tensors: $m, v, x_0, s$ (Algo.~\ref{algorithm:prodigy} Line 11, 12, 17, 18). Consequently, Prodigy can be very memory-intensive when applied to large models with billions of parameters.

\begin{figure}[t]
\begin{algorithm}[H]
    \caption{Prodigy}\label{algorithm:prodigy}
    \begin{algorithmic}[1]
        \Require $t$ (step), $\eta$ (lr, default 1 with cosine annealing), $\theta^i$ (i-th params), $L(\theta)$ (loss function), $\lambda$ (weight decay), $\beta_1, \beta_2$ (betas),  $maximize$, $d_0 > 0$ (default $1e^{-6}$), $x_0$
        
        \Repeat
        \For{$t\gets 1$}
            \If{$maximize$}
                \State $g^i_t \gets -\nabla L(\theta^i_{t-1})$
            \Else
                \State $g^i_t \gets \nabla L(\theta^i_{t-1})$
            \EndIf
            \State
            \State \text{/* do momentum */}
            \If{$t > 1$}
                \State {$m^i_t \gets \beta_1 m^i_{t-1} + \text{$(1-\beta_1)$} d_t g^i_t $}
                \State $v^i_t \gets \beta_2 v^i_{t-1} + (1-\beta_2) d_t^2 (g^i_t)^2 $
                
            \Else
                \State \text{$m^i_t \gets (1-\beta_1) d_t g^i_t $}; $v^i_t \gets (1-\beta_2) d_t^2 (g^i_t)^2 $
                \State $r_{t-1} = 0$; $s_{t-1} = 0$
            \EndIf

            \State{$r_t = \sqrt{\beta_2} r_{t-1} + (1-\sqrt{\beta_2}) \eta d_t^2 \langle g^i_t, x_0-x_t \rangle$}

            \State{$s_t = \sqrt{\beta_2} s_{t-1} + (1-\sqrt{\beta_2}) \eta d_t^2 g^i_t $}
        
            \State{$ \hat{d_{t+1}} = \frac{r_k}{\|s_t\|_1} $}

            \State{$ d_{t+1} = max(d_k, \hat{d_{t+1}})$}
            \State
            
            \State{\text{/* do weight decay */}}
            \State \text{$\theta^i_t \gets \theta^i_{t-1} - \lambda \eta \theta^i_{t-1}$}
            \State
            \State $\theta^i_t \gets \theta^i_{t-1} - \text{$ \frac{\eta d_t}{\sqrt{v^i_t} + d_t \epsilon}m^i_t $} $

        \EndFor
        \Until {epochs end}
    \end{algorithmic}
\end{algorithm}
\end{figure}

\section{Profiling Results on "g-SNR Calculation" Stage}

As discussed in Section~\ref{sec:sup:optimizer_breakdown}, our method requires calculating the scale ratio during the initial step of the optimization process, which we refer to as the "g-SNR Calculation" stage. This procedure may take extra time because it involves calculating both the gradient norm and its standard deviation. As shown in Table~\ref{table:rtx3090_warmup_result} and Table~\ref{table:a100_warmup_result}, although square root operations are generally considered more computationally intensive than standard float addition and multiplication, the time taken for the "g-SNR Calculation" is still relatively small. Since this calculation is performed only once during each training procedure by design, its duration is negligible compared to the overall training iterations. Therefore, our method remains efficient for optimization in the long run.

\begin{table}[!htbp]
\centering
\small 
\setlength{\tabcolsep}{6pt} 
\renewcommand{\arraystretch}{1.3} 
\resizebox{\columnwidth}{!}{
\begin{tabular}{l|l|l|l}
\hline
Model                      & Method           & Iter Times (ms) & g-SNR Calc (ms) \\ \hline
\multirow{2}{*}{ViT-S/16}  & SGDM             & 12.2 $\pm$ 2.9  & 0                 \\
                           & SGD-SaI (ours) & 13.7 $\pm$ 3.8  & 14.5              \\ \hline
\multirow{2}{*}{ViT-H/14}  & SGDM             & 48.6 $\pm$ 7.8  & 0                 \\
                           & SGD-SaI (ours) & 65.5 $\pm$ 10.0 & 43.3              \\ \hline
\multirow{2}{*}{GPT2-1.5B} & SGDM             & 287.4 $\pm$ 0.9 & 0                 \\
                           & SGD-SaI (ours) & 340.1 $\pm$ 1.1 & 267.6             \\ \hline
\end{tabular}
}
\caption{RTX 3090 Profile Results. The results here are based on a single NVIDIA GeForce RTX 3090 GPU. The trials were conducted over 20 iterations, recording the time taken for each optimization step, which is referred to as the "iteration time" column. We compared the time taken for the g-SNR calculation stage and found that it takes an equal amount of time or less than an optimization step. However, since this calculation is only performed once, it is considered tolerable. }
\label{table:rtx3090_warmup_result}
\end{table}

\begin{table}[!htbp]
\centering
\small 
\setlength{\tabcolsep}{6pt} 
\renewcommand{\arraystretch}{1.3} 
\resizebox{\columnwidth}{!}{
\begin{tabular}{l|l|l|l}
\hline
Model                      & Method           & Iter Times (ms) & g-SNR Calc (ms) \\ \hline
\multirow{2}{*}{ViT-S/16}  & SGDM             & 7.1 $\pm$ 0.0   & 0                 \\
                           & SGD-SaI (ours) & 13.0 $\pm$ 0.1  & 13.7              \\ \hline
\multirow{2}{*}{ViT-H/14}  & SGDM             & 51.9 $\pm$ 10.7 & 0                 \\
                           & SGD-SaI (ours) & 63.1 $\pm$ 7.8  & 106.0             \\ \hline
\multirow{2}{*}{GPT2-1.5B} & SGDM             & 353.2 $\pm$ 0.9 & 0                 \\
                           & SGD-SaI (ours) & 392.8 $\pm$ 0.2 & 353.8             \\ \hline
\end{tabular}
}
\caption{A100 PCIe 40GB Profile Results. It follows the same setting as Table~\ref{table:rtx3090_warmup_result}, expect for the GPU type.}
\label{table:a100_warmup_result}
\end{table}

\section{Algorithm Overview}

In this section, we will provide an overview of how SGD-SaI operates. As a non-adaptive gradient method, SGD-SaI calculates the preconditioned scaling factor based on the g-SNR values before applying the first batch of data in the optimization step. These scaling factors vary across different partitions, as they are closely linked to the architectures being utilized. However, once established, they remain constant throughout the entire training process.

While Adam-mini is memory-efficient, its complex partition rules and repetitive local gain recalculation result in significant computational costs. The improvement in throughput compared to Adam(W) is primarily due to the ability to reduce memory usage, allowing for larger batch sizes and enabling more data to be processed in parallel. In contrast, we not only reduce the memory footprint but also eliminate the entire adaptive local gain calculation, achieving a significant breakthrough in both memory and computational efficiency.

\end{document}